\documentclass{bmcart}

\usepackage{amsthm,amsmath,amsfonts}
\usepackage[utf8]{inputenc} 

\startlocaldefs

\usepackage{textcomp,url}
\usepackage[colorlinks,bookmarksopen,bookmarksnumbered,citecolor=red,urlcolor=red]{hyperref}
\usepackage{graphicx}
\usepackage{epic}
\usepackage{eepic}
\usepackage{epsf}
\usepackage{psfrag}
\usepackage{natbib}
\usepackage{wasysym}
\renewcommand{\phi}{\varphi}
\newcommand{\be}{\begin{equation}}
\newcommand{\ee}{\end{equation}}
\newcommand{\R}{{\mathbb R}}
\newcommand{\E}{{\mathbb E}}
\newcommand{\ind}{{\bf 1}}
\newcommand{\sgn}{{\rm sgn}}

\endlocaldefs

\begin{document}

\begin{frontmatter}

\begin{fmbox}
\dochead{Research}

\title{An unambiguous cloudiness index for nonwovens}

\author[
  addressref={itwm},
  email={michael.godehardt@itwm.fraunhofer.de}
]{\inits{M.}\fnm{Michael} \snm{Godehardt}}
\author[
  addressref={itwm},
  email={ali.moghiseh@itwm.fraunhofer.de}
]{\inits{A.}\fnm{Ali} \snm{Moghiseh}}
\author[
  addressref={itwm,tuk},
  email={christine.oetjen@itwm.fraunhofer.de}
]{\inits{C.}\fnm{Christine} \snm{Oetjen}}
\author[
  addressref={hda},
  email={joachim.ohser@h-da.de}
]{\inits{J.}\fnm{Joachim} \snm{Ohser}}
\author[
  addressref={itwm},                   
  corref={itwm},                       
  email={katja.schladitz@itwm.fraunhofer.de}   
]{\inits{K.}\fnm{Katja} \snm{Schladitz}}

\address[id=itwm]{
  \orgdiv{Image Processing Department},       
  \orgname{Fraunhofer ITWM},                  
  \city{Kaiserslautern},                      
  \cny{Germany}                               
}
\address[id=tuk]{
  \orgdiv{Department of Mathematics},       
  \orgname{Technische Universität Kaiserslautern},                  
  \city{Kaiserslautern},                      
  \cny{Germany}                               
}
\address[id=hda]{
  \orgdiv{Department of Mathematics and Natural Sciences},       
  \orgname{University of Applied Sciences Darmstadt},                  
  \city{Darmstadt},                      
  \cny{Germany}                               
}




\end{fmbox}

\begin{abstractbox}

\begin{abstract}
  Cloudiness or formation is a concept routinely used in industry to
  address deviations from homogeneity in nonwovens and papers.
  Measuring a cloudiness index based on image data is a common task in
  industrial quality assurance. The two most popular ways of
  quantifying cloudiness are based on power spectrum or correlation
  function on the one hand or the Laplacian pyramid on the other
  hand. Here, we recall the mathematical basis of the first approach
  comprehensively, derive a cloudiness index, and demonstrate its
  practical estimation. We prove that the Laplacian pyramid as well as
  other quantities characterizing cloudiness like the range of
  interaction and the intensity of small-angle scattering are very
  closely related to the power spectrum. Finally, we show that the
  power spectrum can be measured easily by image analysis methods and 
  carries more information than the alternatives.
\end{abstract}



\end{abstractbox}
%

\end{frontmatter}

\section{Introduction}
Nonwovens, papers, and felts are used in a wide variety of fields,
ranging from filtration of gases or liquids over thermal insulation
and soundproofing to hygiene and sanitary articles, see e.g.\
\cite{durst07}.  They consist of fibers of limited length or
filaments, that are more or less randomly distributed and forming
the solid matter of a macroscopically homogeneous porous
material. 

The macroscopic properties of nonwovens are significantly determined by
the specific weight (i.e.\ the mean weight per unit area, also called
the nominal grammage) and the spatial weight distribution. Moreover, the
nonwoven texture -- also called `nonwoven cloudiness' or `chart
cloudiness' -- is of high practical relevance.  In paper industry, `formation' and
`flocculation' are used as synonyms for `chart cloudiness'.

Cloudiness influences filter properties like flow rate, particle
retention efficiency, wet strength, porosity, dust holding capability,
heat transfer, and sound attenuation \cite{schladitz06}. 
Thus, quantifying cloudiness is
important for industrial quality control as well as for the
development of new filter materials and manufacturing technologies.
Robust estimation of a cloudiness index could help in
optimizing processing parameters.

Holding a sheet of nonwoven up against light yields a visual
impression of cloudiness as the spatial distribution of darker or
brighter regions.  Unfortunately, the term `cloudiness' has not yet
been defined in any industry standard for nonwovens. The industrial
standard \cite{iso4046} on paper, board, pulps, and related terms describes
`formation' only very roughly as `manner in which the
fibers are distributed, disposed, and intermixed to constitute the
paper' and `look-through' as `structural appearance of a sheet of paper
observed in diffuse transmitted light'.

The intensive efforts that have been taken to characterize cloudiness
become evident by the large variety of dedicated publications,
see Section~\ref{sec:conc-cloud-paper} below.  Nevertheless, so far,
no characteristic or index for quantifying cloudiness could be agreed
on. Fixing `cloudiness' in an industry standard seems to be out of
reach. This is due to the wide variety of nonwoven fabrics, the rapid
development of physical and optical testing methods, a wide range of
image analysis methods that are difficult to survey and, finally, the
understandable interest of optical inspection system manufacturers to
keep their specific concepts secret.

Here, we follow
\cite{sara78,norman86,cresson88,provatas96,cherkassky98,chi-ho00,alava06,
  lien06,lehmann13} in using the power spectrum or -- equivalently --
the correlation function to measure cloudiness. The power spectrum and
related mathematical concepts are carefully introduced in
Section~\ref{sec:power-spectrum}. Moreover, the power spectrum of the
modified Bessel correlation function is found to be a flexible
parametric model for the rotation averages of the power spectra of
nonwovens. For the sake of better interpretability, we introduce an
explicit structure model for nonwovens in the subsequent
Section~\ref{sec:model-cloudiness}. Estimation of the suggested cloudiness index
based on images is demonstrated using microscopic images of three
nonwoven samples in Section~\ref{sec:estimation-power-sectrum}.

Finally,  in Section~\ref{sec:other-measures}, we compare our approach 
to a variety of alternatives: the
range of interaction, the Laplacian pyramid, and the intensity of
small-angle scattering. The range
of interaction promoted as homogeneity measure in \cite{xin10,zeng10}
is just the integral of the correlation function.  The Laplacian
pyramid decomposes the image of the structure into `several scales',
evaluates the degree of homogeneity on each scale, and finally derives
an overall degree of homogeneity
\cite{weickert96,weickert99,scholz99}.  Small-angle scattering (SAS)
has been considered as early as in the 1970s, too, to characterize the
cloudiness of paper formation \cite{alsnielsen11}. This is an obvious
choice, as the scatter intensity is closely related to the
power spectrum. We recall this relation in
Section~\ref{sec:small-angle-scatt}, as recent devices make SAS
attractive for industrial quality control.  In general, we contribute
to a better overview by showing in Section~\ref{sec:other-measures}
how closely the quantities listed here are
related. Moreover, we describe under which conditions cloudiness can
be characterized completely.

\section{Concepts of cloudiness of paper and nonwovens}
\label{sec:conc-cloud-paper}
There is a vast variety of publications on characterizing the
cloudiness of nonwovens and formation of paper based on transmission
of diffuse light, where a transmission light table is applied to
ensure homogeneous illumination
\cite{kallmes84,cresson88,cherkassky99,drouin01}. See
\cite{waterhouse91,praast91} for a comprehensive survey on literature
from the late 1980s and \cite{chinga-carrasco09} for more recent
developments. 

Robertson's mean flock size \cite{robertson56} is an intuitive
characteristic for paper formation. However, there is still no
convincing method for segmenting flocks in transmitted light images
available. Practically more useful methods are based on estimating the
variance of the pixel values or, more general, the co-occurrence
matrix of the image data
\cite{yuhara86,cresson88,cresson90,cresson90a}.  In
\cite{pourdeyhimi02}, a `uniformity index' of texture is suggested
based on Poisson statistics for the centers of paper flocks
(objects). This approach seems a bit inconsistent however. On the one
hand, the flock centers cannot be detected
robustly. On the other hand, the index is finally computed based on local area
fraction variation in a binarized image not using the flock centers at all.

In the carefully written monograph \cite{deng94}, an index of
cloudiness is defined as the ratio of two variances: For the nonwoven
sample and a reference model, local grammage is averaged over a square
of edge length 1\,mm. The ratio of the variances of these two
characteristics yields the index.
This approach dates back to earlier works, see e.g.\ \cite{norman74}
frequently cited in recent literature like \cite{sampson09}, too.
Farnood's approach \cite{farnood95,farnood97} is similar, modeling
the fluctuation of the local grammage by a Poisson
shot noise process (i.e.\ a dilution process) of spherical
flocks. In
\cite{ianson03}, Farnood's flocculation
characteristic is derived from the so-called power spectrum, also known as
`power spectral density' or `Bartlett spectrum', of the pattern in
transmission images.

Woven textiles usually feature a periodic pattern. Hence, applying
Fourier methods for quality inspection seems natural, see e.g.\
\cite{wang11,chan:pang2000} and references therein. Periodicity of the pattern
corresponds to characteristic peaks of the power spectrum, and slight
deviations from perfect periodicity can lead to smearing (i.e.\
broadening) of these peaks \cite{ianson95}. Following this line of
thought, transferring these ideas to non-periodic but macroscopically
homogeneous patterns and characterizing nonwoven fabrics with the help
of spectral analysis seems obvious.  In
\cite{sara78,norman86,cresson88,provatas96,cherkassky98,chi-ho00,lien06,lehmann13},
the power spectrum (or equivalently the correlation function) is
suggested to measure cloudiness, see also Section 2.2 in
\cite{alava06}.

Further approaches are based on modeling random structures by Markov
random fields and decomposing the image of the structure into `several
scales', evaluating the degree of homogeneity on each scale and
computing an overall degree of homogeneity. In \cite{scholz99}, this
approach is applied originally on the structure of nonwovens. These
`several scales' are also known as the Laplacian or Gaussian-Laplacian
or Burt-Adelson pyramid for image data
\cite{weickert96,weickert99,scholz99} originally introduced
for image compression \citep{burt83}.

Scharcanski \cite{scharcanski06} suggests a wavelet transform for evaluating sheet
formation and cloudiness of nonwovens.  Replacing the Fourier
transform by a wavelet transform for the analysis of the 
local grammage
in the frequency domain can have computational advantages over the
Fourier transform, especially when recursive methods are applied for
calculation. Wavelets (in particular Haar wavelets) are very well
suited for analyzing piece-wise constant functions like microscopic
images of the microstructure of multi-phase materials. The local
grammage of nonwoven however is not piece-wise constant.
Nevertheless, there are wavelet transforms for a
variety of purposes and the choice of one of them influences the
estimation of the spectral density. For example, the hybrid method
presented in \cite{maldonado07} -- a combination of a discrete wavelet
transform and Gabor filter banks -- aims at the segmentation of clouds
in a pre-specified size range.

Mathematical modeling of nonwoven structures on a mesoscale also yields a
deeper understanding of the phenomenon of cloudiness
\cite{cresson88,cherkassky98,antoine00,gregersen00,provatas00,sampson09}.
Model parameters can be interpreted as characteristics of cloudiness
(i.\,e. the uniformity index or the degree of homogeneity of the
pattern), given they can easily be estimated from image data. In
\cite{lehmann13}, Gaussian random fields are used for modeling
paper structure and a Fourier approach is applied for computing these
characteristics, following suggestions from \cite{xu96,lien06}.


\section{Power spectrum}
\label{sec:power-spectrum}
Obviously, the accuracy of quantities estimated by optical methods 
always depends on the imaging conditions. Here, we introduce the 
quantities independent of the imaging parameters. Their limits are 
discussed later.

Assume the structure of a nonwoven to be captured in a transmitted
light image using a point, line, or area detector. In this case, the
local grammage of the nonwoven and the detected signal can be expected
to be closely related. In fact, often, these two are not really
distinguished.
Nevertheless, we want to emphasize the difference: Let $\R^2$ be the
projection plane orthogonal to the optical axis, and let
$\gamma$ be the linear mass attenuation coefficient of the solid
matter. Then the intensity $I(x)$ of the transmitted light at a point
$x=(x_1,x_2)$ in the plane $\R^2$ can be related to the local grammage
$w(x)$ by Lambert-Beer's law $I(x)=I_0\exp(-\gamma w(x))$, where $I_0$
is the initial intensity of the applied light. Thus, the local
grammage is 
\be
\label{eqGrammage} 
w(x)=\frac{1}{\gamma}\bigl(\ln I_0 - \ln I(x)\bigr),
\qquad x\in\R^2.
\ee 
See \cite{vandenakker49,mcdonald86,lien06} for the computation of the
local grammage from the absorption of visible light. The use of
$\beta$- and soft X-radiation is suggested in
\cite{komppa88,kajanto89,komppa96,scanp09}, and \cite{farrington88},
respectively, and the application of electron beam transmission
for estimating local grammage is studied in \cite{keller98}.  Finally,
the influence of the choice of radiation on transmittance is
investigated for nonwovens in \cite{boeckerman92} and for paper in
\cite{norman76,bergeron88}.

\begin{figure}[htb]
\begin{center}
\includegraphics[width=.8\textwidth]{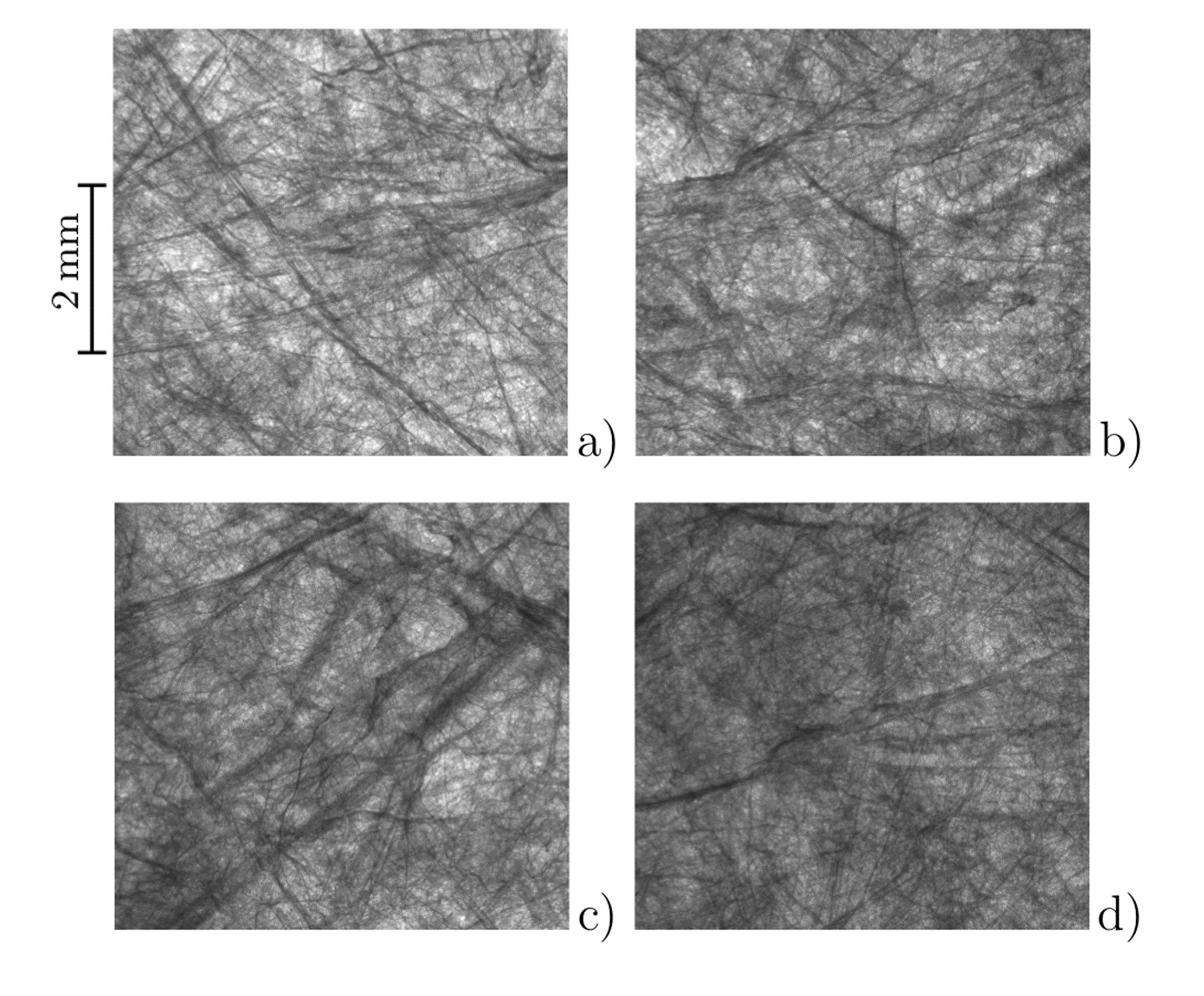}
\end{center}
\caption{Four microscopic images of a nonwoven layer from a face mask. Several
  fields of view taken in transmission mode, $2\,048\times 2\,048$
  pixels of size 3.367\,\textmu m. The total area of the four images is 190.2\,mm$^2$.}
\label{figOpt1_1x-sandler}
\end{figure}

Nonwovens feature a random, macroscopically
homogeneous structure, i.e. the distribution of the structure is invariant with
respect to translations in $\R^2$. Hence, the local grammage $w(x)$ is a
random function on $\R^2$ that is macroscopically homogeneous as
well. We denote by $\bar{w}=\E w(x)$ and
$\sigma_w^2=\E\bigl(w(x)-\bar{w}\bigr)^2$ the mean grammage
(also called the `nominal grammage' or the `base weight') and the
variance, respectively, where $\E$ denotes the expectation.
Furthermore, let 
\be
\label{eq:normalized-grammage}
f(x)=\bigl(w(x)-\bar{w}\bigr)/\sigma_w
\ee 
be the normalized local grammage. Then the correlation function $k(x)$ of
$w(x)$ is simply given by
\[ k(x) = \E\bigl(f(y)f(y+x)\bigr),\qquad x,y\in\R^2,\] 
where the right-hand side of this equation is independent of the position
$y$ because of the macroscopic homogeneity of $f(x)$. Finally, the
power spectrum $\hat{k}(\xi)$ of $f(x)$ is defined as the
2-dimensional (2D) Fourier transform of the correlation function
$k(x)$, 
\be
\label{eqFourierT}
\hat{k}(\xi)=\frac{1}{2\pi}\int\limits_{\R^2}k(x)\,\exp(-ix\xi)\,dx,
\qquad
\xi\in\R^2.
\ee 
Vice versa, $k(x)$ is the Fourier co-transform of
$\hat{k}(\xi)$, 
\be
\label{eqFourierCT}
k(x)=\frac{1}{2\pi}\int\limits_{\R^2}\hat{k}(\xi)\,\exp(ix\xi)\,d\xi,
\qquad
x\in\R^2.
\ee 
The variable $\xi=(\xi_1,\xi_2)$ is known as the circular
frequency. Note that Equation~\eqref{eqFourierT} is motivated by the
projection-slice theorem of the Fourier transform stating that the 2D
Fourier transform of the projection of a 3D structure is a slice of
the 3D Fourier transform of the structure
\cite{bracewell12}. Analogously, the 2D power spectrum $\hat{k}(\xi)$
of $f(x)$ is a slice of the 3D power spectrum of the 3D nonwoven
structure.

Often, nonwoven structures are not only macroscopically homogeneous,
but also isotropic, i.e.\ the distribution of the structure is
invariant with respect to rotations in $\R^2$. Then the correlation
function $k(x)$ and the power spectrum $\hat{k}(\xi)$ are isotropic as
well, which means that they depend only on the radial coordinates
$r=\|x\|=\sqrt{\vphantom{X^2}x_1^2+x_2^2}$ and
$\varrho=\|\xi\|=\sqrt{\vphantom{Xi^2}\xi_1^2+\xi_2^2}$, respectively. So there are
functions $k_1(r)$ and $\hat{k}_1(\varrho)$ with $k_1(\|x\|)=k(x)$,
$x\in\R^2$ and $\hat{k}_1(\|\xi\|)=\hat{k}(\xi)$, $\xi\in\R^2$,
respectively, related to each other by the Bessel transform
\be\label{eqBessel}
\hat{k}_1(\varrho)=\int\limits_0^\infty k_1(r)\,r\,J_0(r\varrho)\,dr,\qquad \varrho\geq 0,
\ee
where $J_0(r)$ is the Bessel function of the 1st kind and of order
0. See e.\,g. \cite{katznelson04} for a sound introduction to the Fourier
transform and related topics. The Bessel transform, 
also known as the Hankel transform, is inverse to
itself, i.\,e. it is an involution.
Consider for example the modified Bessel correlation function
\be
k_1(r)=\frac{(\lambda r)^\nu\,K_\nu(\lambda
  r)}{2^{\nu-1}\Gamma(\nu)},\qquad r\ge 0, 
\label{eq:Bessel-corr}
\ee
with the parameters $\lambda>0$, $\nu>0$ and the modified Bessel
function of the 2nd kind $K_\nu(r)$ of order $\nu$.
Its Bessel transform is
\be
\label{eqBessPow} 
\hat{k}_1(\varrho)=\frac{2\nu
  \lambda^{2\nu}}{(\lambda^2+\varrho^2)^{\nu+1}},\qquad\varrho\ge 0,
\ee
see \cite{abrahamsen85}.

\begin{figure}[htb]
\begin{center}
\includegraphics[width=0.75\textwidth]{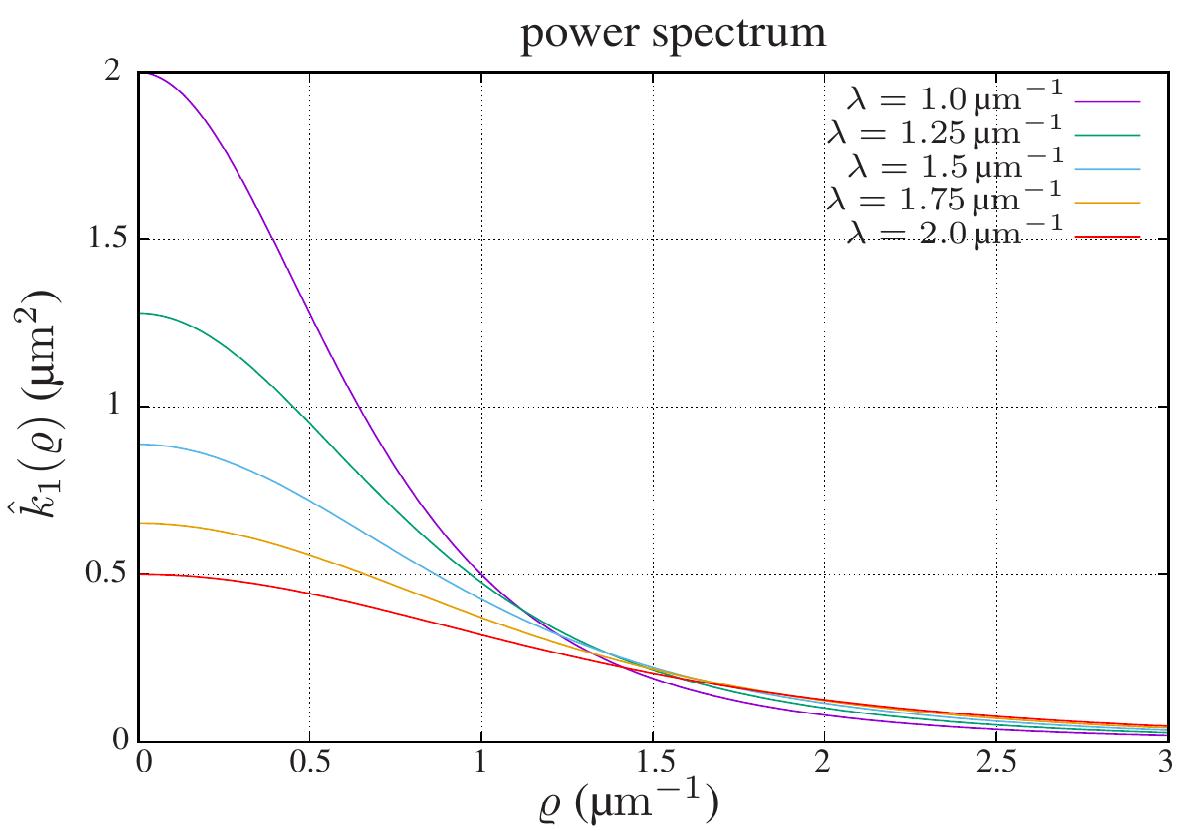}
\end{center}
\caption{Graphs of the power spectrum $\hat{k}_1(\varrho)$ given by
  \eqref{eqBessPow} for constant $\nu=1$ and varying
  $\lambda$.}
\label{figModBess}
\end{figure}

Graphs of the above power spectra are shown in
Figure~\ref{figModBess}. The power of low frequencies decreases with
increasing $\lambda$. From Equation~\eqref{eqFourierCT} one
immediately gets 
\be
\label{eqPowerConservationLaw}
1=k(0)=\frac{1}{2\pi}\int\limits_{\R^2}\hat{k}(\xi)\,d\xi=\int\limits_0^\infty
\varrho\,\hat{k}_1(\varrho)\,d\varrho\ ,
\ee 
i.\,e. the integral of the power spectrum is constant. 
From this kind of `energy conservation law' it follows that decreasing
the power of low frequencies (i.\,e. decreasing cloudiness) leads to
increasing the power of higher frequencies. For high frequencies
($\varrho\gg\lambda$), the power spectrum $\hat{k}_1(\varrho)$ given
by \eqref{eqBessPow} decays as fast as $\varrho^{-2(\nu+1)}$.

It is often suggested to characterize cloudiness by the variance
$\sigma_w^2$ of the local grammage $w(x)$ or its distribution
function. 
We would like to stress that this is impossible.  A simple scaling
$w(cx)$ of the local weight of a nonwoven with a constant $c>1$
results in the nonwoven appearing more homogeneous due to the strongly
decreasing power spectrum. In fact, because of the inverse scaling law
of the Fourier transform, the power spectrum of $w(cx)$ is
$\hat{k}(\xi/c)/c^2$, and $\hat{k}(\xi/c)/c^2<\hat{k}(\xi)$ for small
frequencies, whereas
variance and distribution of the local grammage are not affected by
the scaling.

\section{Modeling cloudiness}
\label{sec:model-cloudiness}
The function defined by~\eqref{eqBessPow} is certainly sufficiently
flexible to be fit to measured power spectra. However, the parameters
$\lambda$ and $\nu$ are hard to interpret. In order to
clarify the relation of nonwoven structure and power spectrum,
we model the nonwoven structure
and derive a formula for the power spectrum $\hat{k}_1(\varrho)$.

\begin{figure}[htb]
\begin{center}
\includegraphics[width=.75\textwidth]{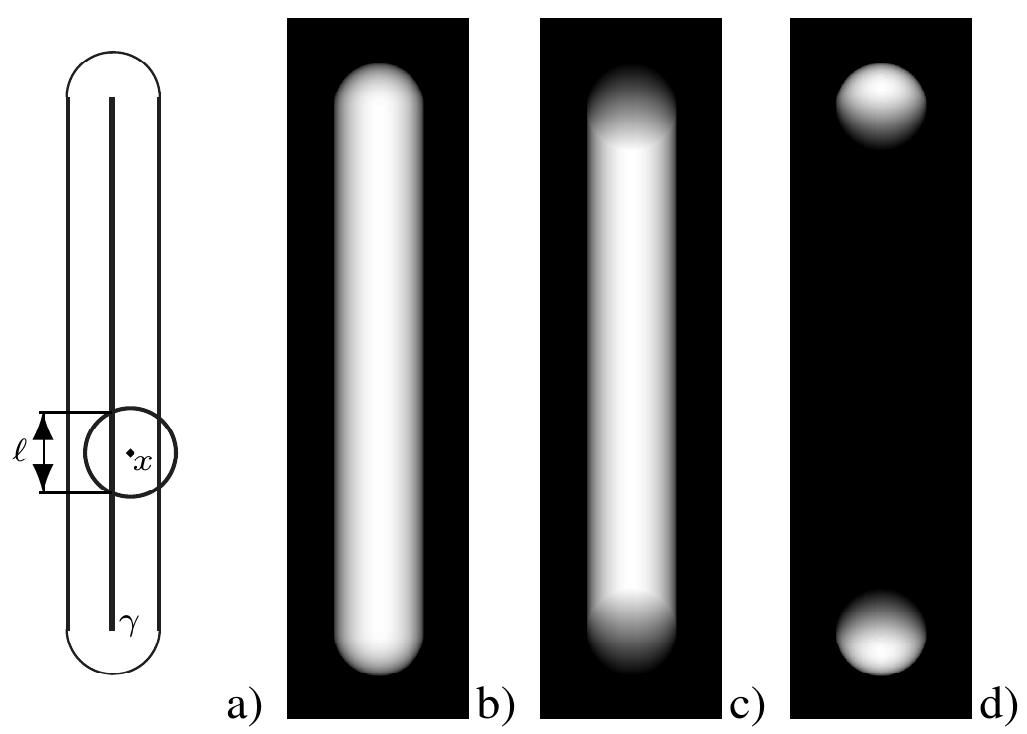}
\end{center}
\caption{Modeling the local grammage of a straight fiber: 
a) the core $\gamma$ and the contour of the fiber model $\gamma\oplus
B_R$, 
b) the projection of the fiber model on the plane $\R^2$, 
c) the core convolved with the filter mask $\kappa(x)$, 
d) the dif\-fer\-en\-ce between projection and filtering. 
The filter response in $x$ is the length $\ell$ of the core $\gamma$ 
intersected with the circle divided by the area of the circle.
}
\label{figFiberModel}
\end{figure}

We first model the local grammage $w(x)$ of a nonwoven. 
Let $\phi(s)$, $0\leq s\leq 1$ be a parametric function in the
projection plane $\R^2$. Then the set $\gamma=\{\phi(s), 0\leq s\leq
1\}$ is a curve in $\R^2$. We assume $\gamma$ to be of finite length. 
Its dilation $C=\gamma\oplus B_R$ with the ball $B_R$ of radius $R$
yields a filled tube of circular cross-section, see
Figure~\ref{figFiberModel}a. Now $C$ can serve as a 
model of a curved fiber of constant thickness $2R$.
We assume $C$ to be morphologically regular. This requires in
particular the curvature of the core $\gamma$ to be less than $1/(2R)$. 
The orthogonal projection $p(x)$ of $C$ on $\R^2$ is the chord length
of $C$ at $x\in\R^2$, i.\,e. the length of the intersection of $C$
with the straight line orthogonal to the projection plane $\R^2$ and
hitting the point $x$, see Figure~\ref{figFiberModel}b.

\begin{figure}[htb]
\begin{center}
\includegraphics[width=0.75\textwidth]{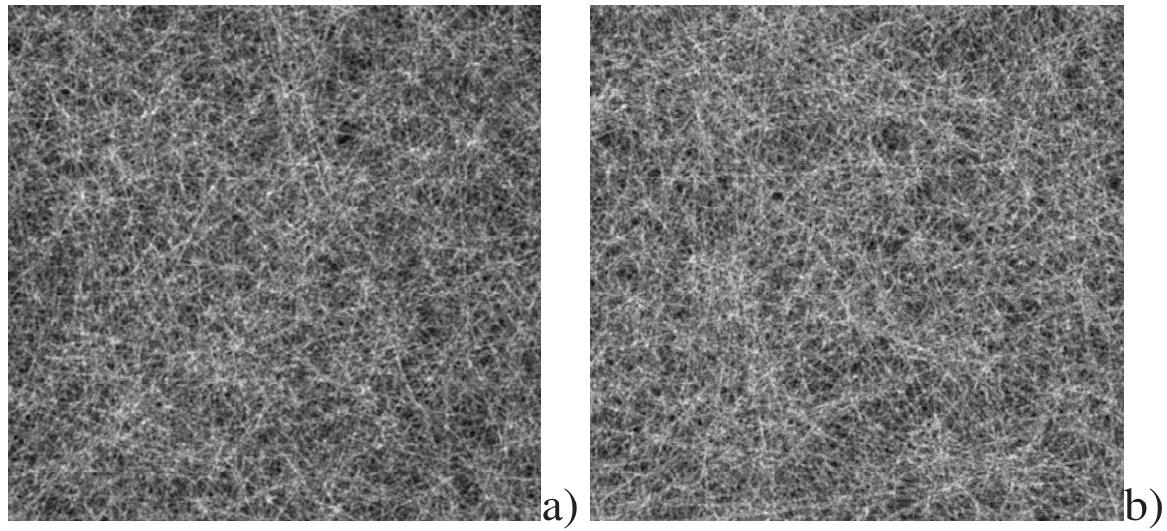}
\end{center}
\caption{Realizations of random fields generated by segments of
  exponentially distributed lengths, $1\,024\times 1\,024$ pixels of
  size 4.0\,\textmu m, thickness of the tubes 6.0\,\textmu m: 
  a) mean length 0.5\,mm, b) mean length 1\,mm.}
\label{figGRF}
\end{figure}

Now we consider a macroscopically homogeneous and isotropic random
system of fiber cores. That is, parametric functions $\phi_j(s)$, $0\leq s\leq 1$,
$j=1,2,\ldots$, where the points $\phi_j(0)$, $j=1,2,\ldots$ form a
macroscopically homogeneous Poisson point field. Finally, the local
grammage $w(x)$ is modeled by the macroscopically homogeneous and
isotropic random field $w_{\rm m}(x)$ defined as the sum of the
orthogonal projections $p_j(x)$ of the fiber models $C_j$ with the
cores $\gamma_j=\{\phi_i(s), 0\leq s\leq 1\}$ times the specific mass
density $\rho$ of the fiber material,
\[ w_{\rm m}(x) = \rho\sum\limits_{j=1}^\infty p_j(x),\qquad
  x\in\R^2.\]
See Figure~\ref{figGRF} for examples of random fields $w_{\rm m}(x)$
generated by segments $\gamma_j$ with directions uniformly
distributed in $[0,\pi)$ and exponentially distributed lengths.

Let $\bar{w}_{\rm m}$ and $\sigma_{\rm m}$ be the mean and the
standard deviation, respectively, of the random field $w_{\rm
  m}(x)$. Then the normalized local grammage
$f_{\rm m}(x)=(w_{\rm m}(x)-\bar{w}_{\rm m})/\sigma_{\rm m}$ converges
to a Gaussian random field of mean 0 and standard deviation 1 as the
mean number of curves per unit area $N_A$ goes to infinity due to the
central limit theorem \cite{adler81}. That is, for sufficiently large
$N_A$, the random field $w_{\rm m}(x)$ is characterized exclusively by
the mean $\bar{w}_{\rm m}$, the standard deviation $\sigma_{\rm m}$,
and the power spectrum $\hat{k}_{1,{\rm m}}(\varrho)$. As a
consequence, we can assert that the power spectrum
$\hat{k}_1(\varrho)$ of $w(x)$ characterizes cloudiness uniquely if
$w_{\rm m}(x)$ is a suitable model for the local grammage $w(x)$ of
the nonwoven.

Aiming at a closed formula for the power spectrum
$\hat{k}_{1,{\rm m}}(\varrho)$ of the model $w_{\rm m}(x)$, we now
replace the
projections $p_j(x)$ of the sphero-cylinders $C_j$ by the cores
$\gamma_j$ convolved with a suitable
filter kernel $\kappa(x)$. Then $p_j(x)$ is approximated by the
filter response in the point $x\in\R^2$, see
Figure~\ref{figFiberModel}c. Practically, this is achieved by filtering the Bresenham line
of the segment $\gamma_i$ with the mask of an isotropic mean value filter:
\[ \kappa(x) =\left\{\begin{array}{ll}\displaystyle{\frac{1}{\pi
                       R^2}},\quad & 
                                     \|x\|\leq R\\&\vspace{-0.3 cm}\\ 0,
                                   &\mbox{otherwise}
                     \end{array}\right.,
\qquad x\in\R^2.\]
 
\begin{figure}[htb]
\begin{center}
\includegraphics[width=0.75\textwidth]{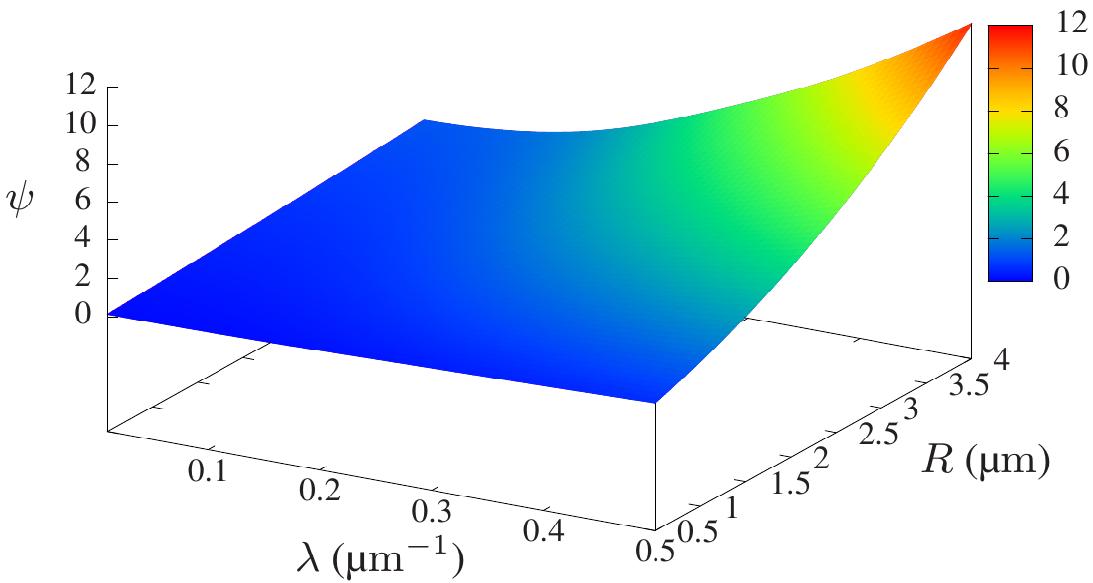}
\end{center}
\caption{The correction factor $\psi(\lambda,R)$ from \eqref{eq:psi} used in \eqref{eqGRFPower}.}
\label{figPsi}
\end{figure}

The Bessel transform of the radial function $\kappa_1(r)$ of $\kappa(x)$ is given by
\[ \hat{\kappa}_1(\varrho)=\frac{J_1(R\varrho)}{\pi R\varrho},\qquad\varrho>0,\]
with the Bessel function $J_1(r)$ of the first kind and of order 1
\cite{katznelson04}. The pair correlation function of the straight line system is 
\[ {\rm pcf}(r) = 1+\frac{\lambda}{\pi N_Ar}\,\exp(-\lambda r),\qquad r>0,\]
see \cite{chiu13}. Its Bessel transform is 
\[ \widehat{\rm pcf}(\varrho)=
   \delta(x)+\frac{1}{\pi
     N_A}\frac{\lambda}{\sqrt{\lambda^2+\varrho^2}},
   \qquad\varrho\geq 0,\]
with the Dirac function $\delta(x)$, see \cite{lehmann13}.
Hence, the Fourier convolution theorem yields that the power spectrum
$\hat{k}_{1,{\rm m}}(\varrho)$ of $w_{\rm m}(x)$ generated by a system
of straight fibers of radius $R$ can be approximated as
\be
\label{eqGRFPower}
\hat{k}_{1,{\rm m}}(\varrho) \approx 
\frac{\psi(\lambda,R)}{\lambda\sqrt{\lambda^2+\varrho^2}}
     \cdot\frac{J_1^2(R\varrho)}{R^2\varrho^2},\qquad\varrho\geq 0,
\ee
where $\psi(\lambda,R)>0$ given by
\be
\psi^{-1}(\lambda, R) =
\frac{2\pi}{\lambda
  R^2}\int\limits_0^\infty\frac{J_1^2(R\varrho)}{\varrho^2\sqrt{\lambda^2+\varrho^2}}\,d\varrho 
\label{eq:psi}
\ee
is a correction factor ensuring that $\hat{k}_{1,{\rm
    m}}(\varrho)$ is actually the Bessel transform of a correlation
function $k_{1,{\rm m}}(r)$, in particular $k_{1,{\rm m}}(0)=1$. 
For high frequencies, the right-hand side of \eqref{eqGRFPower}
resembles qualitatively the square of sine or cosine waves whose
amplitude decays as $\varrho^{-4}$, and whose first factor is as
\eqref{eqBessPow} for $\nu=-\frac{1}{2}$. Unfortunately, until now,
there is no closed form known for $\psi(\lambda,R)$. For numerical
values see Figure~\ref{figPsi}.

\begin{figure}[htb]
\begin{center}
\includegraphics[width=0.75\textwidth]{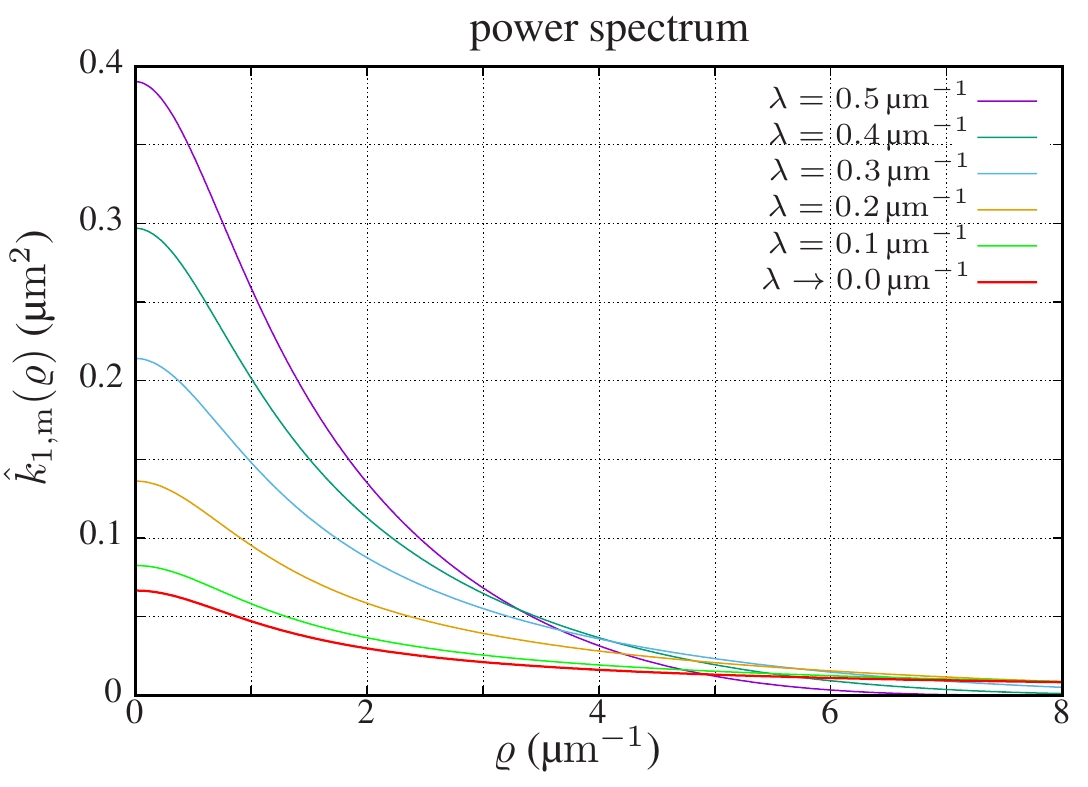}
\end{center}
\caption{Graphs of the power spectrum $\hat{k}_{1,{\rm m}}(\varrho)$
  given by \eqref{eqGRFPower} for constant radius
  $R=1\,$\textmu m and varying $\lambda$.} 
\label{figPSModel}
\end{figure}

The parameter $\lambda$ of the exponential distribution is the inverse
mean segment length, i.\,e. $1/\lambda+2R$ is the mean fiber length.
Figure~\ref{figPSModel} shows the power spectrum
$\hat{k}_{1,{\rm m}}(\varrho)$ given by
Equation~\eqref{eqGRFPower}. The
power of low frequencies (assigning small chart
cloudiness) decreases with increasing mean fiber length. The limiting case of
infinitely long fibers ($\lambda=0$) yields the smallest cloudiness. The right side of the
approximation in \eqref{eqGRFPower} converges to the left side as
$\lambda\to 0$ (infinite fiber length, filaments), since the
projection and filter response differ only at the fiber ends, see
Figure~\ref{figFiberModel}d.

Regime changes for the rate of the decay of the correlation function
may be observed for fixed fiber length. However, this does not apply
here, as fiber lengths are randomly distributed.

\section{Estimation of the power spectrum}
\label{sec:estimation-power-sectrum}

The accuracy of quantities estimated by optical methods always depends
on the stimulating light (intensity, wavelength, coherence), the size
of the investigated sample (i.e. the window size), and the type of
detector (pixel size, modulation transfer function).  These
metrological parameters have to be adjusted to allow for reliable
measurements in any case. Here, we assume sufficient quality of the
images throughout. We discuss in this section however details on how
to estimate the power spectrum from image data unambiguously.

The pixel value $g(x)$ of a gray-tone image at the pixel position $x$
does never match exactly the intensity $I(x)$ of the transmitted
light. The intensity $I(x)$ can be deduced from $g(x)$ only if the
imaging conditions are known, including the photo-detector spectral
response, its dark current density, and the sensor conversion
gain. Thus, estimating mean grammage $\bar{w}$ and
  standard deviation $\sigma_w$ objectively requires tedious
  calibration.  However, in order to quantify cloudiness, only the
power spectrum $\hat{k}(\xi)$ of the relative local grammage $f(x)$
\eqref{eq:normalized-grammage} has to be estimated. 

In the following
we assume $g(x)>0$ for all pixel positions $x$ of the image. In
practice this means that the illumination at the microscope
(intensity, exposure time) must be chosen such that the gray-tone is
positive almost everywhere. Whenever $g(x)$ and $I(x)$ are
approximately proportional, then $f(x)$ can be estimated
from 
\be
\label{eqGrammage1} 
f(x)\approx\frac{1}{\sigma}\bigl(\mu-\ln
g(x)\bigr),\qquad x\in\R^2,
\ee 
cf. Equation~\eqref{eqGrammage}, where $\mu$
and $\sigma$ are the mean and the standard deviation of the function
$\ln g(x)$ for all $x$, respectively.

\subsection{Estimation based on a finite field of view}
\label{sec:window}

In this section, we deal with the practical implications of the fact,
that the imaged field of view is always finite. 

Let $W\subset\R^2$ be an image frame or field of view of area
$A$. Let $\ind_W(x)$ denote the associated indicator function,
$\ind_W(x)=1$ for $x\in W$, and $\ind_W(x)=0$ otherwise. The
windowed function $f_W(x)=f(x)\ind_W(x)$, i.\,e.\ the image
$f_W(x)$ of $f(x)$ in $W$, is square integrable. Thus the 
correlation of $f_W(x)$ with itself
is well defined and can be written as the convolution of
$f_W(x)$ with the reflection $f_W^*(x) = f_W(-x)$ of $f_W(x)$ at the
origin,
\[ \bigl(f_W*f_W^*\bigr)(x)=\int\limits_{\R^2} f_W(y)f_W(y-x)\,dy \]
for all $x$ in the interior $W^\circ$ of $W$. Here $*$ denotes convolution.
Then the correlation function $k(x)$ of $f(x)$ can be rewritten as
\[ k(x) = \frac{\bigl(f_W*f_W^*\bigr)(x)}{c_W(x)},\qquad x\in W^\circ,\]
where $c_W(x)=\bigl(\ind_W*\ind_W^*\bigr)(x)$ is the so-called 
window function. If $x$ is small compared to the window size, then $c_W(x)$ is 
approximately the frame area, $c_W(x)\approx c_W(0)=A$. 
The use of a Fast Fourier Transform (FFT) 
induces additionally a periodic extension of
$f_W(x)$, which together with the Fourier convolution theorem 
leads to the approximation
\be
\label{eqEstPow} 
\hat{k}(\xi)\approx\frac{2\pi \E|\hat{f}_W(\xi)|^2}{A},
\qquad\|\xi\|\gg 0,
\ee
for estimating the power spectrum $\hat{k}(\xi)$ by image analysis. 
See \cite{koch03,ohser05,lehmann13} for exact estimation of $\hat{k}(\xi)$.

\subsection{Averaging over several images}
\label{sec:pixelwise-mean}

The limited field of view implied by the imaging method can be
compensated by taking images of several disjoint fields of view.

The linearity of the Fourier transform and the macroscopic homogeneity
of nonwovens result in the following important fact: For several
images $f_{W,j}(x)$, $j=1,\ldots,m$ taken at disjoint positions of
$W$, the power spectrum of the mean
$f_W(x)=\frac{1}{m}\sum_{j=1}^m f_{W,j}(x)$ coincides with the power
spectra of the individual images. That means, we can first pixelwise
average the $f_{W,j}(x)$ and then estimate the power spectrum of the
mean.

As a consequence, instead of $m$ time-consuming FFTs,
only one FFT is needed for estimating the power spectrum of
$f_W(x)$. Thus, the number $m$ of images can be increased arbitrarily
while the computing time remains almost constant. 

\subsection{Application to microscopic images of nonwovens}
\label{sec:material}

We consider three nonwoven samples. Our first example is a
polypropylene nonwoven used as one layer in face masks. Specimen FL1
is produced by Sandler AG, Schwarzenbach/Saale, Germany. It has a
nominal grammage of 100\,gm$^{-2}$, thickness of 0.98\,mm at 0.05\,Ncm$^{-1}$,
initial pressure drop of 163\,Pa, and air permeability of
98\,Lm$^{-2}$s$^{-1}$ at 200\,Pa according to the industrial standard
\cite{nwsp19}. About 35\,\% of the fibers have a thickness $d$ smaller
than 1.25\,\textmu m, 33\,\% with
1.25\,\textmu m$\leq d<$2.5\,\textmu m, 29\,\% with
2.5\,\textmu m$\leq d<$5\,\textmu m, and 3\,\% with
$d\geq$ 5\,\textmu m.

The structure is imaged in transmission mode using the Leica stereo
light microscope MZ16. The detailed image conditions are:
planapochromatic $1\times$-objective, numerical aperture 0.14, working
distance 55\,mm; light source
Phlox-LedRGB-BL-$100\times100$-S-Q-IR-24V, white, exposure time 2\,ms;
detector Basler acA4112-8gm with $4\,096\times 3\,000$ pixels; sensor
IMX304 CMOS, GigE, mono.

\begin{figure}[htb]
\begin{center}
\includegraphics[width=.75\textwidth]{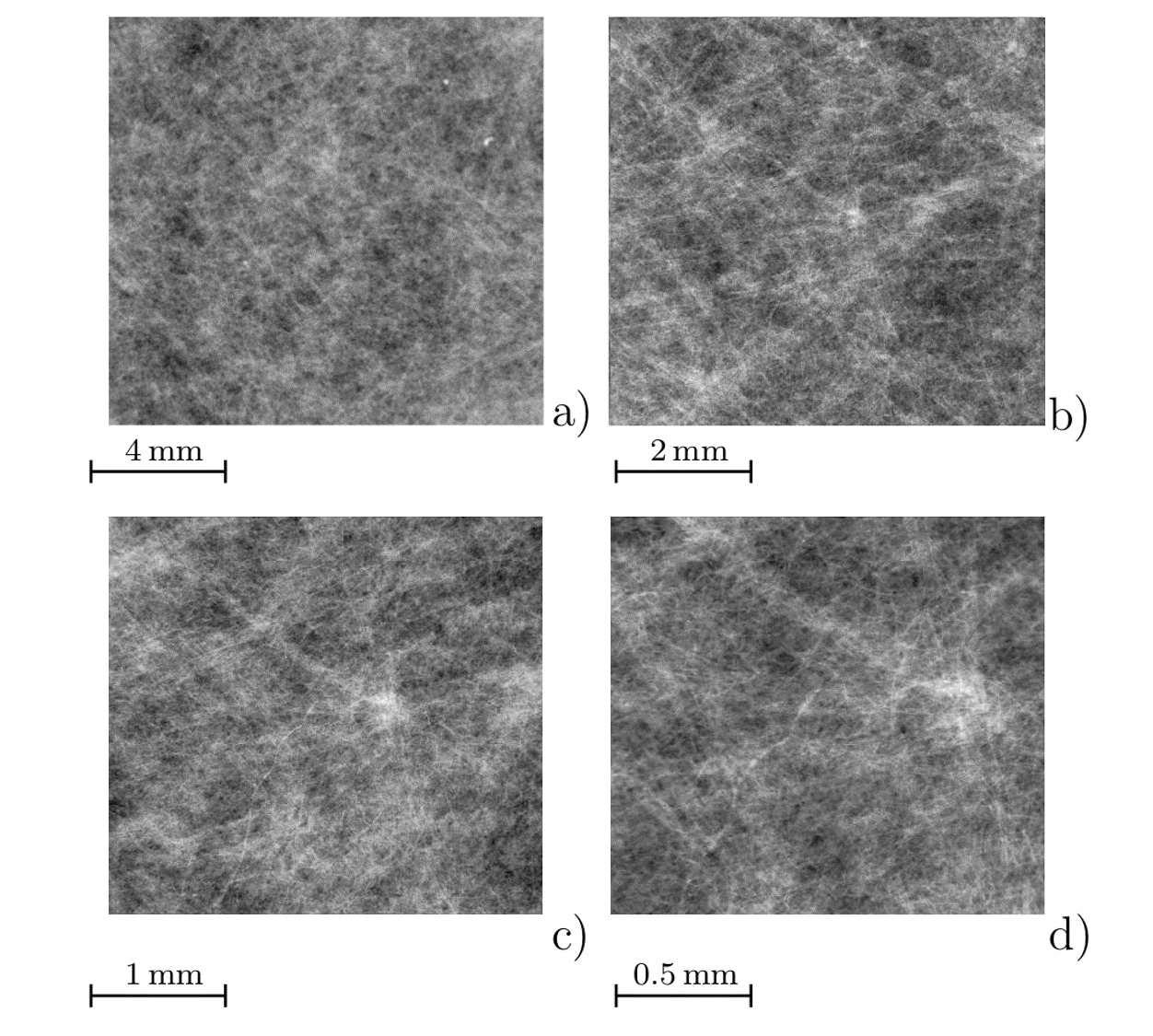}
\end{center}
\caption{Pixelwise mean of the normalized local grammages of the
  nonwoven FL1 for four lateral resolutions. 
  The images consist of $2\,048\times 2\,048$ pixels:
a) pixel size $6.734\,$\textmu m,
b) pixel size $3.367\,$\textmu m,
c) pixel size $1.684\,$\textmu m,
d) pixel size $0.842\,$\textmu m.
}
\label{figOpt1}
\end{figure}

Figure~\ref{figOpt1_1x-sandler} shows $m=4$ gray-tone images of the nonwoven
taken at the same lateral resolution ($2\,048\times 2\,048$ out of
totally $4\,096\times 3\,000$ pixels of size $3.367\,$\textmu
  m). Figure~\ref{figOpt1}b shows the pixelwise mean $f_W(x)$ of the
$f_{W,j}(x)$ obtained from the four images shown in
Figure~\ref{figOpt1_1x-sandler}.

\begin{figure}[htb]
\begin{center}
\includegraphics[width=0.75\textwidth]{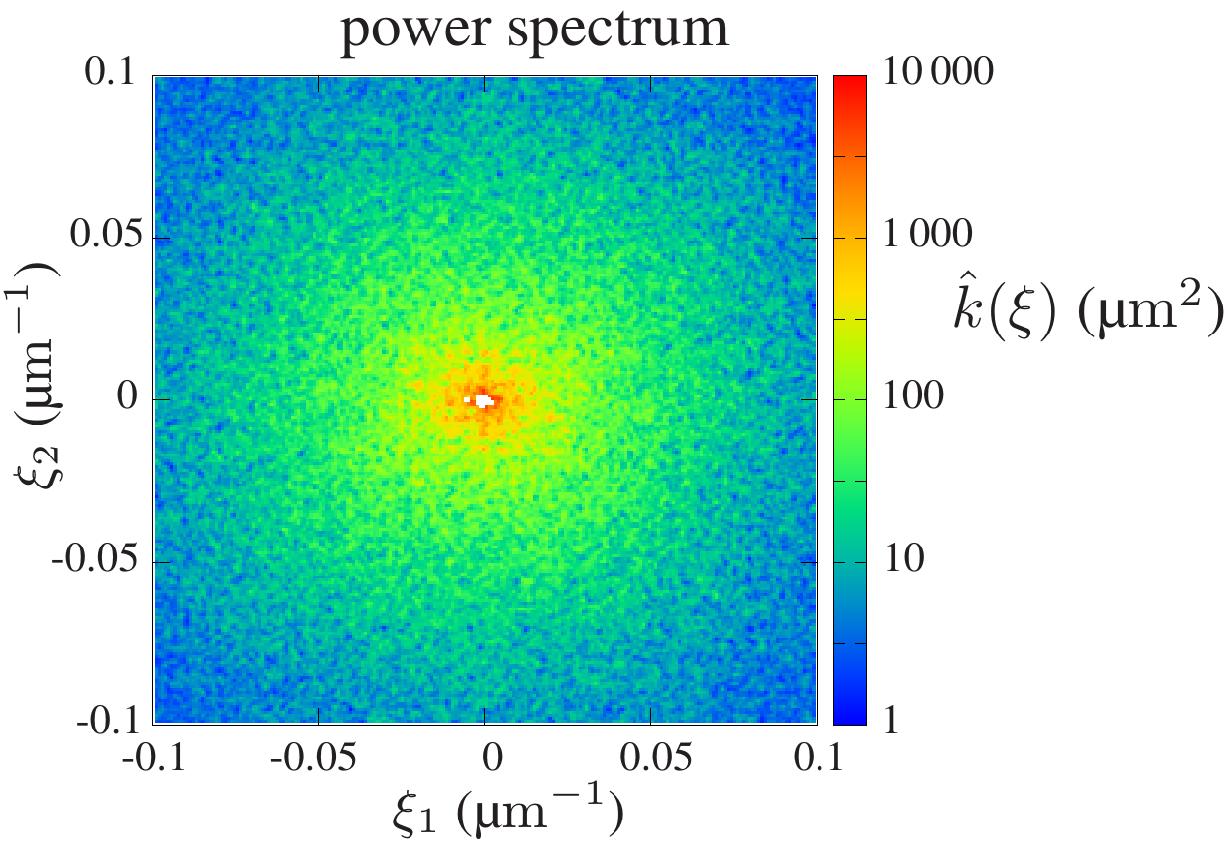}
\end{center}
\caption{Power spectrum $\hat{k}(\xi)$ of the nonwoven
  FL1, estimated based on the image from Figure~\ref{figOpt1}a,
  as a density plot depending on  $\xi=(\xi_1,\xi_2)$ and in logarithmic scale.}
\label{figProbePow1}
\end{figure}

Figure~\ref{figProbePow1} shows the power spectrum $\hat{k}(\xi)$ of
the normalized local grammage of the nonwoven FL1 from
Figure~\ref{figOpt1}a. The isotropy of FL1 is passed
on to $\hat{k}(\xi)$, which obviously is invariant with respect to
rotations about the origin. Thus the radial function
$\hat{k}_1(\varrho)$ contains the same information as
$\hat{k}(\xi)$. It is therefore obvious to determine
$\hat{k}_1(\varrho)$ as the rotation average of the estimate of
$\hat{k}(\xi)$, which also reduces the
statistical error.

\begin{figure}[htb]
\begin{center}
\includegraphics[width=0.95\textwidth]{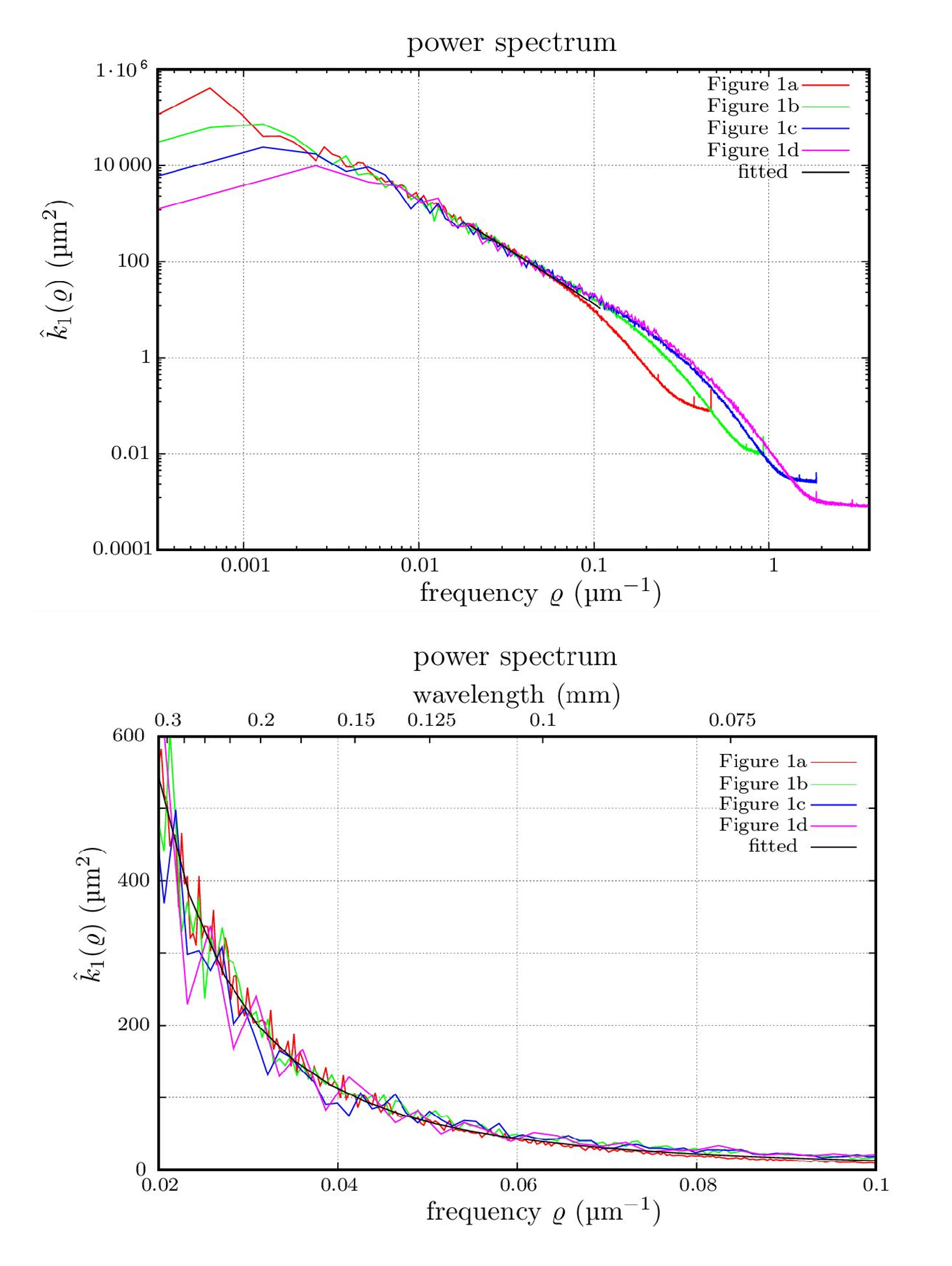}\\
\end{center}
\caption{Estimates of the power spectrum of the nonwoven FL1 from the
  images in Figures~\ref{figOpt1}a-d and the
  function from \eqref{eqBessPow} fitted to the power spectrum
  of Figure~\ref{figOpt1}a with $\lambda=6.4\,$mm$^{-1}$ and
  $\nu=0.231$. Top: logarithmic scale. Bottom: zoomed into sub-range, 
  linear scale.}
\label{figPowOpt1}
\end{figure}

Figure~\ref{figPowOpt1} shows the power spectrum $\hat{k}_1(\varrho)$  
of the nonwoven FL1 estimated based on images at varying lateral
resolution (Figures~\ref{figOpt1}a to \ref{figOpt1}d). 
Additionally, the model function~\eqref{eqBessPow} is
fit to the estimated power spectrum. 

Total area and lateral resolution have to be chosen such that
$\hat{k}_1(\varrho)$ can be estimated in a given spectral band with a
sufficiently small total error. The statistical error increases for
decreasing $\varrho$, and it diverges as $\varrho\downarrow 0$. For
higher frequencies the systematic error increases as fine details of
the nonwoven structure are not resolved anymore. Thus, the lateral
resolution and the size of one field of view determine the eligible
spectral band as the frequency $\varrho$ has to be larger than the
inverse length of the image diagonal and smaller than twice the
inverse pixel size.

If this assumption holds, then the statistical estimation error of
$\hat{k}_1(\varrho)$ decreases with increasing total area covered by
the images. It is thus desirable to extend the field of view as wide
as possible. Alternatively, several fields of view can be averaged as
shown in Section~\ref{sec:pixelwise-mean} above.
 
Finally, note that estimates of the power spectrum of $f(x)$ are 
largely independent of small deviations from the proportionality of 
the functions $g(x)$ and $I(x)$.

Our samples FL2 and FL3 are polyester spunbond
nonwovens as applied in sanitary products. FL2 has a nominal 
grammage of 28\,gm$^{-2}$, a fiber thickness of
10\,\textmu m and a permeability of 112\,Lm$^{-2}$s$^{-1}$ at
200\,Pa. For FL3, the nominal grammage is 35\,gm$^{-2}$,
the fiber thickness is 25\,\textmu m, and the permeability at
200\,Pa is 315\,Lm$^{-2}$s$^{-1}$. 

\begin{figure}[htb]
\begin{center}
\includegraphics[width=.75\textwidth]{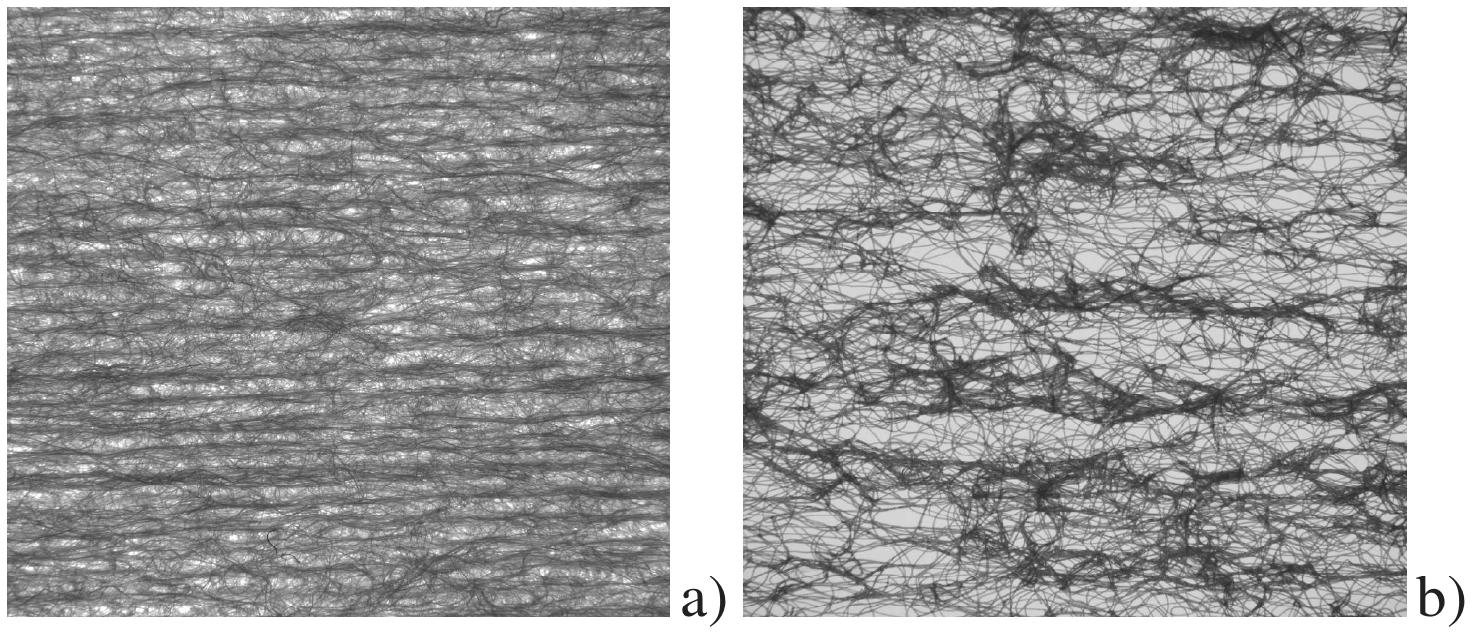}
\end{center}
\caption{Images of the anisotropic nonwovens a) FL2 and b) FL3,
  $2\,048\times 2\,048$ pixels of size 6.734\,\textmu m. The image
  area is 190.2\,mm$^2$.}
\label{figProbe12Img}
\end{figure}

\begin{figure}[htb]
\begin{center}
\includegraphics[width=.75\textwidth]{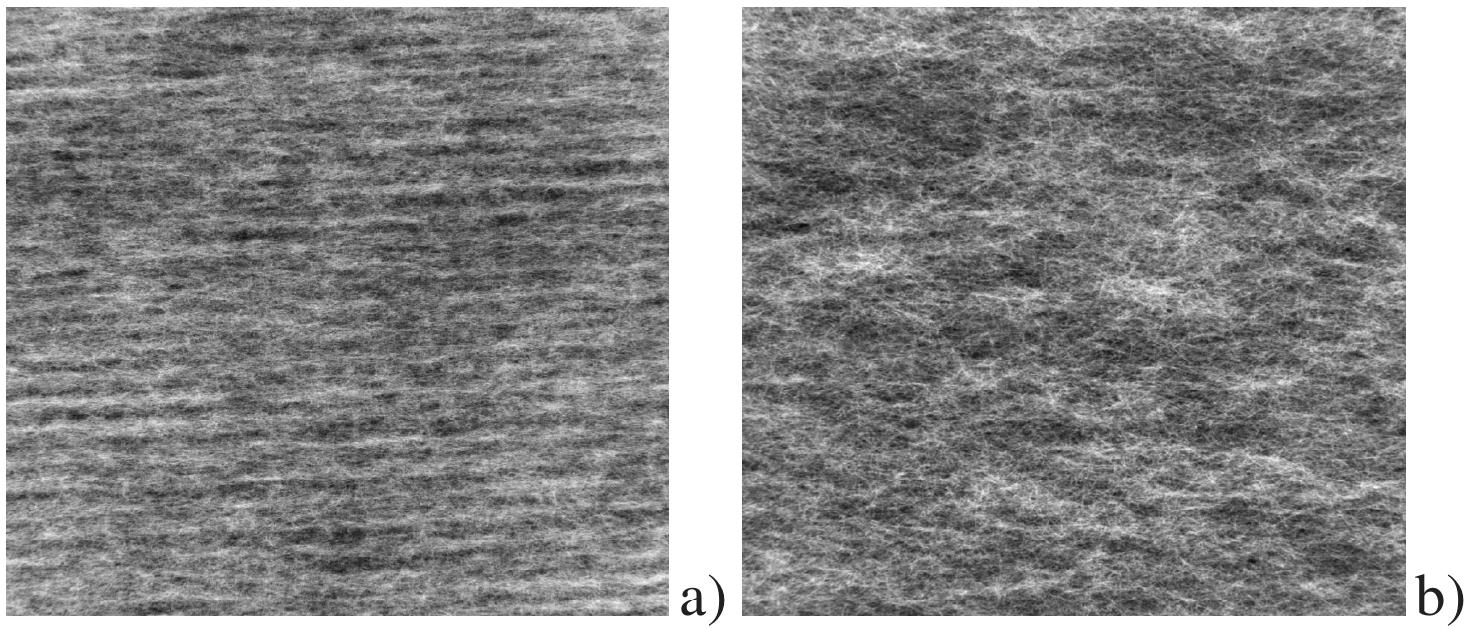}
\end{center}
\caption{Pixelwise mean of the normalized local grammages of
  anisotropic nonwovens a) FL2 and b) FL3, $2\,048\times 2\,048$
  pixels of size 6.734\,\textmu m. The mean was computed from
  $m=10$ separate images of a total area 1\,902.0\,mm$^2$.}
\label{figProbe12}
\end{figure}

The microscopic images are acquired
under the same conditions as for FL1 with the exposure times being the
only difference: 8\,ms for FL2 and 3\,ms for FL3. See
Figures~\ref{figProbe12Img}a and \ref{figProbe12Img}b for example
images. The pixelwise means shown in Figures~\ref{figProbe12}a and
\ref{figProbe12}b are calculated from $m=10$ transmission images.

\begin{figure}[htb]
\begin{center}
\includegraphics[width=.75\textwidth]{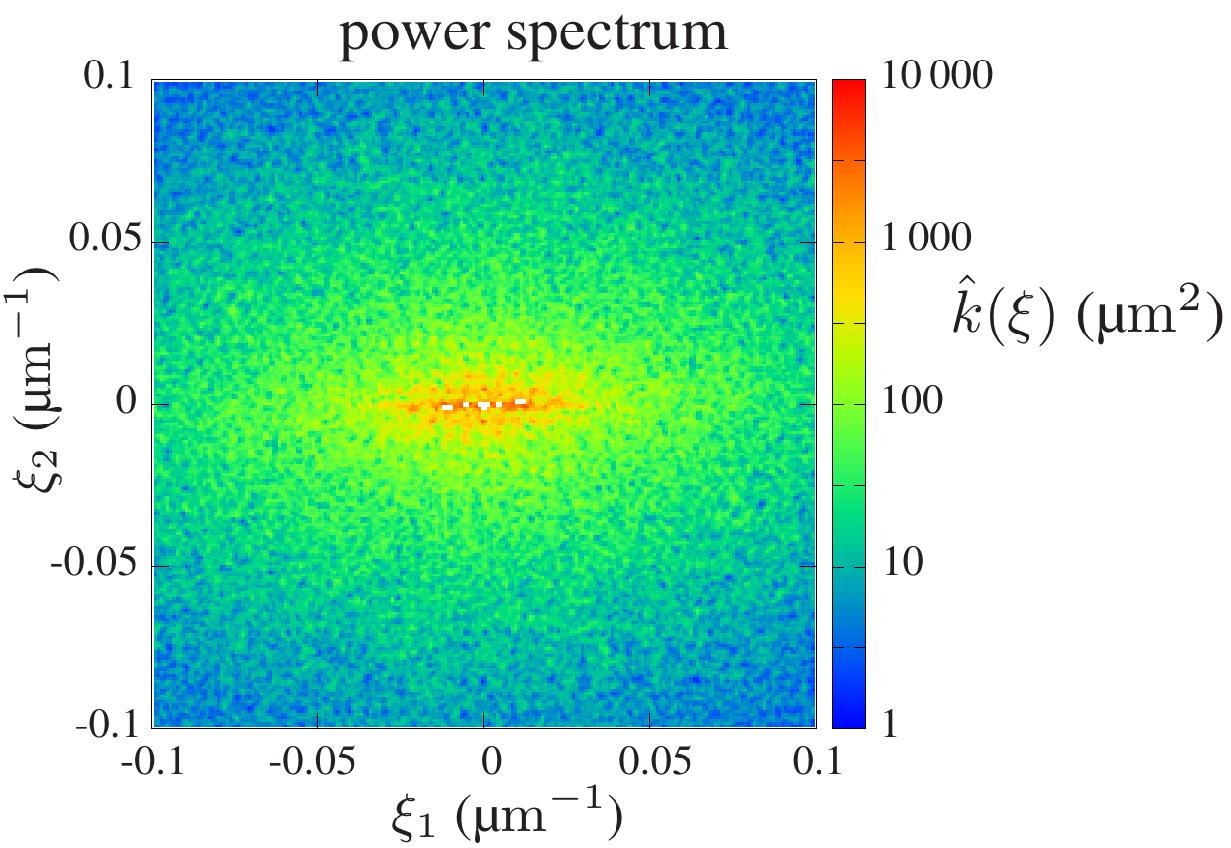}
\end{center}
\caption{Estimate of the power spectrum $\hat{k}(\xi)$ of FL2 based on
  the image Figure~\ref{figProbe12}a.}
\label{figProbePow2}
\end{figure}

\begin{figure}[htb]
\begin{center}
\includegraphics[width=0.75\textwidth]{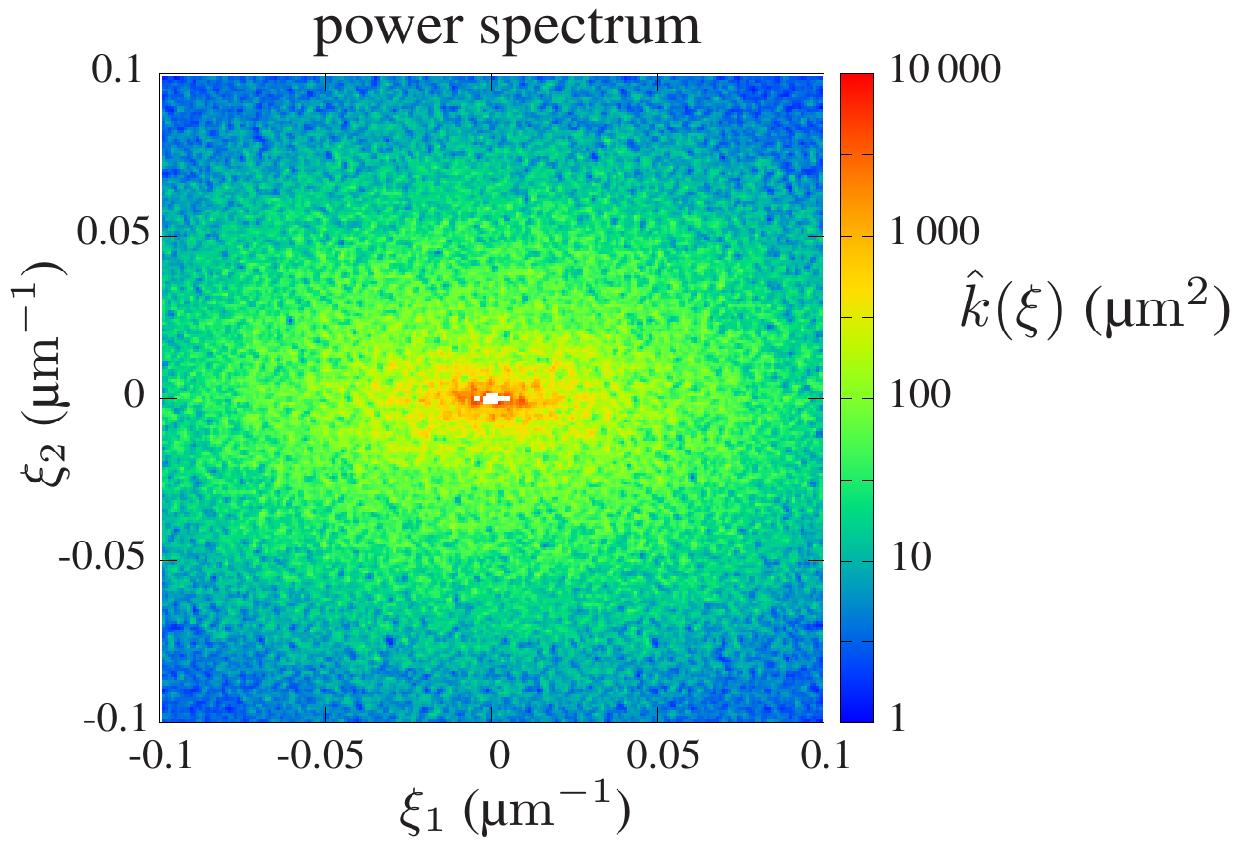}
\end{center}
\caption{Estimate of the power spectrum $\hat{k}(\xi)$ of FL3 based on
  the image Figure~\ref{figProbe12}b.}
\label{figProbePow3}
\end{figure}

\begin{figure}[htb]
\begin{center}
\includegraphics[width=0.75\textwidth]{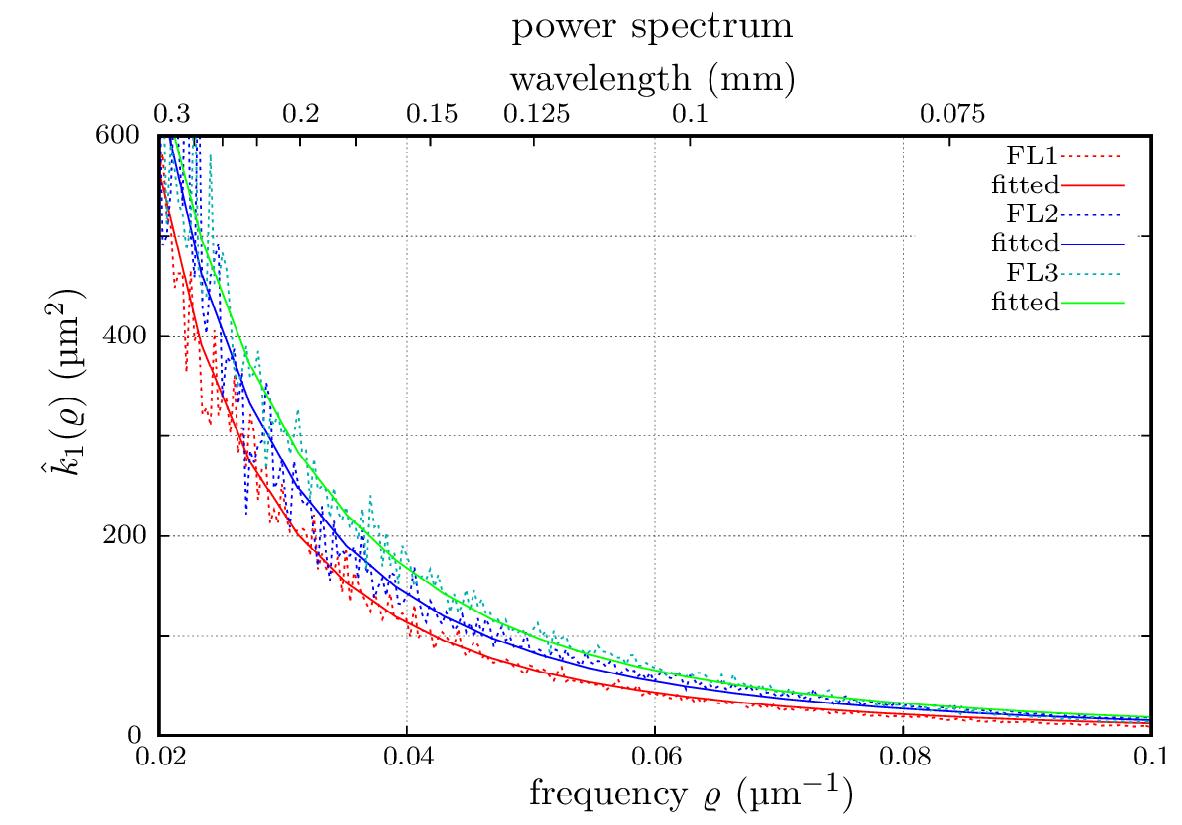}
\end{center}
\caption{Estimates of the power spectrum of the nonwovens FL1, FL2,
  and FL3 from the images in Figures~\ref{figOpt1}a,
  \ref{figProbe12}a and \ref{figProbe12}b, respectively, and 
  the function from \eqref{eqBessPow} fitted to the estimated 
  power spectra. Linear scale.}
\label{figProbePow12}
\end{figure}

Both nonwovens feature a significant degree of anisotropy. That means,
the power spectrum $\hat{k}(\xi)$ is anisotropic too, see
Figures~\ref{figProbePow2} and \ref{figProbePow3}. The rotation
mean $\hat{k}_1(\varrho)$ of the power spectrum can nevertheless be
computed, see Figure~\ref{figProbePow12}. The estimate of
$\hat{k}_1(\varrho)$ for sample FL1 for the same lateral resolution is
included in this figure for comparison. See Table~\ref{tabParam} for
the parameters $\lambda$ and $\nu$ of the adapted model function from Equation~\eqref{eqBessPow}.

\subsection{Cloudiness index}
\label{sec:cloudiness-index}

We now derive the cloudiness index CLI as the total power in a medium
frequency band $[\varrho_0,\varrho_1]$ for $0<\varrho_0<\varrho_1$,
\be
\label{eqCLI} 
{\rm CLI} =
\int\limits_{\varrho_0}^{\varrho_1}\varrho\,\hat{k}_1(\varrho)\,d\varrho.
\ee
Equation~\eqref{eqPowerConservationLaw} yields that $0\leq{\rm
  CLI}\leq 1$. Thus, the CLI can be given in percent. 

The thus defined cloudiness index depends solely on the chosen
frequency band which is determined to a high extent by the imaging
conditions as explained in the previous Section~\ref{sec:material}.  
The lower bound $\varrho_0$ is bounded by the inverse length of the
diagonal of the imaged field of view. The upper bound $\varrho_1$ is 
restricted by the lateral
resolution, that is the doubled inverse pixel size. 
Decreasing $\varrho_0$ or increasing $\varrho_1$ requires
enlarging the field of view or increasing the resolution,
respectively.  
Moreover, an expansion of the range does not
necessarily lead to a better assessment of cloudiness due to the power
conservation law. For industrial quality control, this range has to be
specified according to the application case.

For application example FL1, we use
$\varrho_0=0.02\,$\textmu m$^{-1}$ and
$\varrho_1=0.10\,$\textmu m$^{-1}$, corresponding to wavelengths between
62.8\,\textmu m and 314.2\,\textmu m, and obtain a
cloudiness of ${\rm CLI}=25.7\,\%$. See Table~\ref{tabParam} for the
corresponding results for examples FL2 and FL3.
\begin{table}[htb]
\caption{Estimates of the parameters $\lambda$ and $\nu$ for the
  fitted function \eqref{eqBessPow}, the cloudiness index CLI, and the
  range of interaction RI. The specified error bounds are derived as
  the standard deviation of the mean of the estimates based on the
  individual images. Total sample area is $A$.
}
\label{tabParam} 
\begin{center}
\renewcommand{\arraystretch}{1.3}
\begin{tabular}{|c|r|r||r|r|r|r|}
\hline
spe- & $m$ & $A$     & $\lambda$~~~~~ & $\nu$~~~~~ & CLI~ & RI~~~~ \\
cimen & & (mm$^2$) & (mm$^{-1}$) &  & (\%)~ & (mm$^2$) \\
\hline\hline
FL1 & 14 & 2\,663& $6.400$~ & $0.231$ & $29.7$ & $0.071$ \\ 
    & & & $\pm0.420$~   & $\pm0.015$ & $\pm 0.8$ & $\pm0.015$\\\hline 
FL2 & 10 & 1\,902 & $11.410$~ & $0.226$ & $35.4$ & $0.022$ \\ 
    & & & $\pm0.415$~ & $\pm0.005$ & $\pm0.4$ & $\pm0.002$ \\\hline 
FL3 & 10 & 1\,902 & $15.000$~ & $0.253$ & $39.1$ & $0.014$ \\ 
    & & & $\pm0.251$~ & $\pm0.005$ & $\pm0.5$ & $\pm0.001$ \\ 
\hline
\end{tabular}
\end{center}
\end{table}
Obviously, computing the cloudiness index ${\rm CLI}$ from the
rotation mean $\hat{k}_1(\varrho)$ removes the information on
anisotropy of a nonwoven. A cloudiness index including directional
information could be derived by considering the power spectrum
$\hat{k}(\xi)$ as a function of the polar coordinates $r=\|\xi\|$ and
$\varphi=\arctan\frac{\xi_2}{\xi_1}$ for $\xi_1\neq 0$ and
$\varphi=\pi\,\sgn\,\xi_2$, otherwise, where $\sgn\,\xi_2$ is the sign
of $\xi_2$. Restricting the angle $\varphi$ to a sub-range can then
reveal directional dependency of the structure. For example, for
sample FL2 from Figure~\ref{figProbe12Img}, the cloudiness index is
$17.1\,\%$ for the sector $|\varphi|\leq\frac{\pi}{4}$ (i.\,e. for
directions $\varphi$ close to the processing direction of the
nonwoven) and $45.5\,\%$ for $|\varphi|>\frac{\pi}{4}$.  

Alternatives are e.\,g. the orthogonal projection of the power
spectrum $\hat{k}(\xi)$ onto the x- and y-axis, respectively, or
fitting a 2D function generalizing \eqref{eqBessPow} to
$\hat{k}(\xi)$. In this case, the parameters $\lambda$ and $\nu$ depend on
the angle $\varphi$. Characterization of anisotropy
is however subject of further research.

In inspection systems for industrial quality control, line cameras are
often used to scan the material continuously in the production line,
see e.\,g.\ \cite{isra21,rycobel21}. The focus is on defect
detection. However, cloudiness of nonwovens and formation of paper can
be measured with such systems, too. The 1D correlation function of the
camera signal is a linear section through the 2D correlation function
$k(x)$ through the origin and in line direction. In the isotropic
case, this 1D correlation function coincides with the rotation mean
$k_1(r)$. For the power spectrum, however, this is not true. The 1D
power spectrum of the camera signal obtained with the 1D Fourier
transform is, in the isotropic case, the cosine transform of $k_1(r)$
but not the Bessel transform.

\section{Relation to other quantities characterizing homogeneity}
\label{sec:other-measures}

Many measures of structural homogeneity or deviation from it have been
suggested, see Section~\ref{sec:conc-cloud-paper}. Here, we shed light
on how they are interrelated with particular focus on our CLI based on
the power spectrum and on the Laplacian pyramid as promoted in
\cite{weickert96,weickert99,scholz99}.

First, we recall that the correlation function
$k(x)$ and the power spectrum $\hat{k}(\xi)$ can be converted into
each other by the Fourier transform. That is, these two functions carry
the same information about the local grammage $w(x)$. Therefore it
seems a matter of taste, whether the cloudiness of a nonwoven or
a paper is characterized based on $k(x) $ or
$\hat{k}(\xi)$. 

However, there are tangible advantages of
characterizing cloudiness based on $\hat{k}(\xi)$. On the one hand, cloudiness
is the total power of a medium frequency band. On the other hand,
there are various approaches to estimate $\hat{k}(\xi)$
directly by scattering. Formally, 
to estimate the correlation function $k(x)$ effectively, 
one has to go via frequency space anyway. This involves two time-consuming
Fourier transforms, while only one suffices for
estimating $\hat{k}(\xi)$.

\subsection{The range of interaction}
\label{sec:range-interaction}
In \cite{xin10} and \cite{zeng10} the range of interaction ${\rm RI}$, i.e.
integral of the correlation function, 
\be
\label{eqRI1}  
{\rm RI}=\int\limits_{\R^2}k(x)\,dx = 2\pi\int\limits_0^\infty
rk_1(r)\,dr,
\ee
is suggested as a cloudiness index for nonwovens, see also
\cite{dodson18}. Analogously to
Equation~\eqref{eqPowerConservationLaw}, it 
follows from Equation~\eqref{eqFourierT} that RI is up to a constant 
the power spectrum at 0,
\be
\label{eqRI2} 
{\rm RI} = 2\pi\hat{k}(0) = 2\pi\hat{k}_1(0). 
\ee
For example, the range of interaction for the modified Bessel
correlation \eqref{eq:Bessel-corr} is
\[{\rm RI} = \frac{4\pi\nu}{\lambda^2}.\]
From the parameters of the fitted function in Figure~\ref{figPowOpt1},
we easily obtain the estimate ${\rm RI}=0.046\,$mm$^2$ for the
nonwoven FL1, see also Table~\ref{tabParam} for the other examples. 

Estimating the range of the correlation is apparently very simple as
demonstrated in this example. However, the errors should be taken into
account. Unavoidable large errors in estimating the correlation
function $k_1(r)$ for $r$ near the maximum width $d_{\rm max}$ of the
window $W$ induce large errors in RI when using Equation~\eqref{eqRI1}
as the basis for estimating RI. Moreover, for $r\geq d_{\rm max}$,
$k_1(r)$ cannot be estimated at all. 
This effect can also be seen in
Equation~\eqref{eqRI2}. Direct estimation of $\hat{k}(0)$ is impossible.
Determining $\hat{k}(0)$ by extrapolating a fitted function instead
bears the risk of inducing another error due to inadequate fitting. 
Finally, it is not clear, whether RI really captures the cloudiness of
a nonwoven:  For very small $\varrho$, the power spectrum
$\hat{k}_1(\varrho)$ represents nothing but a gradient in the nonwoven
structure or a shading of the image e.\,g.~due to uneven illumination
of the field of view.

\subsection{Small-angle scattering}
\label{sec:small-angle-scatt}

Small-angle scattering has a long tradition to measure cloudiness, in
particular for paper. In the past, its practical use was hampered by the high
operating expenses. Currently, this obstacle weakens however with
cheaper and more robust hardware rendering small-angle scattering more attractive.
In this section, we outline briefly how to estimate the power spectrum 
by small-angle scattering of
partially coherent light. 
In theses experiments, the scattering intensity $I_{\rm s}$ is
collected depending on the so-called scattering angle $2\theta$, where
$\theta$ is the Bragg angle \cite{alsnielsen11}. 

Let the light have wave length $\lambda$. The scattering intensity is a
function of the magnitude $q=(4\pi\sin\theta)/\lambda$ 
of the momentum transfer
scattering vector, $I_{\rm s}=I_{\rm s}(q)$. 
For spatially isotropic structures, the scattering intensity
$I_{\rm s}(q)$ is, up to a multiplicative constant $c$, the sine
transformation of the spatial correlation seen as a function of the
radial coordinate \cite{klein96,lindner02}. For small-angle X-ray
scattering (SAXS), the factor $c$ can be interpreted physically, see
e.\,g.\ \cite{alsnielsen11}.

For flat structures like thin nonwovens, the orthogonal projection of
the spatial correlation function onto the xy-plane is approximately
the same as the correlation function of the projection $w(x)$ of the
structure. As a consequence, $q\approx\varrho$ and
$c\cdot I_{\rm s}(q)$ is equivalent to the power spectrum
$\hat{k}_1(\varrho)$. That means, $c\cdot I_{\rm s}(q)$ is the Bessel
transform of $k_1(r)$.

SAXS covers a sample area too small to characterize the cloudiness of
nonwovens. This disadvantage is easily overcome by modern scanning
methods like scanning small-angle X-ray scattering (sSAXS,
\cite{gourrier07}) enabling fast scanning of larger sample areas. The
high exerimental effort of SAXS and sSAXS compared to optical methods
limits their use however to nonwovens made of completely opaque
fibers.

Attempts to use small-angle scattering of visible light to determine 
paper formation, see e.\,g.~\cite{helmer02},  did not succeed because 
of the strong absorption. 

\subsection{Laplacian pyramid}
\label{sec:laplace-pyramid}
\begin{figure}[htb]
\begin{center}
\includegraphics[width=0.75\textwidth]{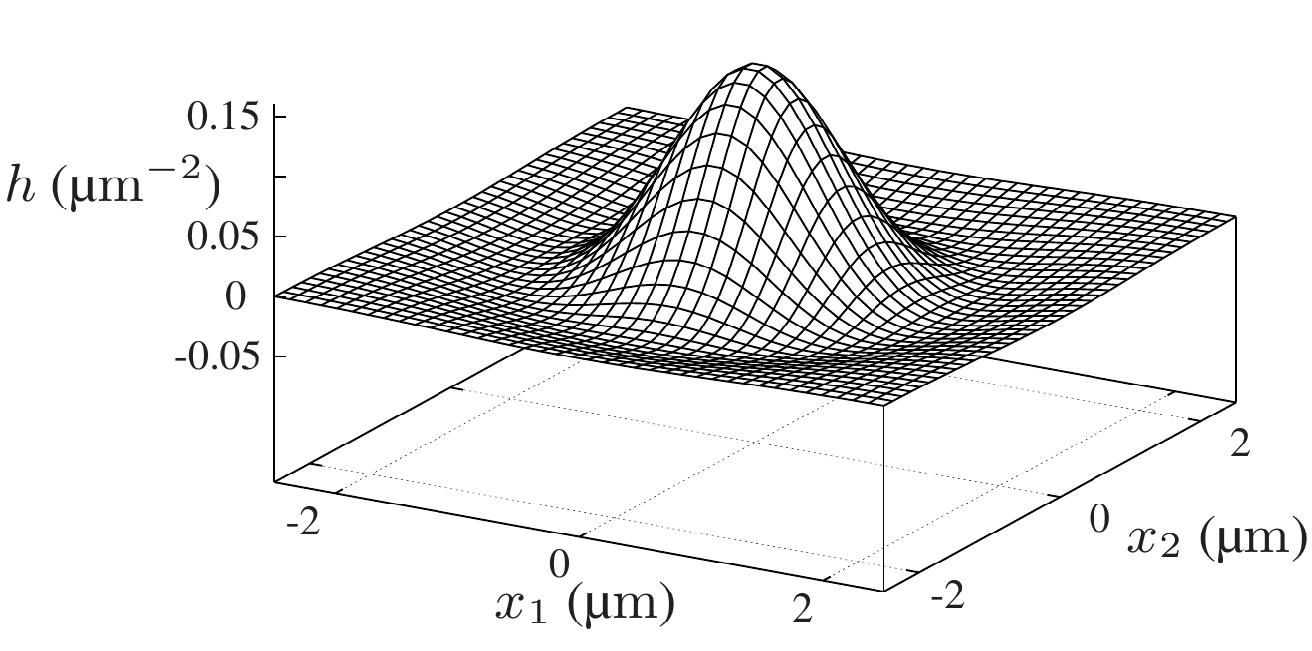}
\end{center}
\caption{Filter mask $h(x)$ of the DoG filter
  with $\sigma_1=\frac{1}{\sqrt{2}}\,$\textmu m and $\sigma_2=1\,$\textmu m.}
\label{figDoG1}
\end{figure}

Characterizing cloudiness of nonwovens by the Laplacian pyramid is based 
on filtering images with differences 
\[ h(x)= g_{\sigma_1}(x)-g_{\sigma_2}(x),\qquad x\in\R^2,\ \sigma_1<\sigma_2,\]
of Gaussian functions given by
\[ g_\sigma(x)=\frac{1}{2\pi\sigma^2}\,\exp\left(-\frac{\|x\|^2}{2\sigma^2}\right),\qquad x\in\R^2.\]
In electrical engineering, 1D versions of these filters are called
band pass filters. The function $h(x)$ is known as the mask of a 
difference of Gaussian (DoG) filter, see Figure~\ref{figDoG1} for an example.

In \cite{weickert96,weickert99} a sequence
$h_j(x)=g_{\sigma_j}(x)-g_{\sigma_{j+1}}(x)$ 
of DoG filters with $\sigma_j=2^{\frac{j-1}{2}}$ is applied to the image
$f_W(x)$. The filter response $\bigl(f_W*h_j\bigr)(x)$ is evaluated by
the mean square of the pixel values ${\rm MSP}\!_j$. In our notation, this is
the ${\mathcal L}_2$ norm of the function $\bigl(f_W*h_j\bigr)(x)$ 
over the window area,
\[ {\rm MSP}\!_j=\frac{\big\|f_W*h_j\big\|^2}{A}=
\frac{1}{A}\int\limits_{\R^2} \bigl|\bigl(f_W*h_j\bigr)(x)\bigr|^2\,dx\]
for integer $j$. In \cite{scholz99}, a weighted sum of the ${\rm
  MSP}\!_j$ is suggested as cloudiness index for nonwovens. The
weights are chosen empirically, driven by customer's demands.

\begin{figure}[htb]
\begin{center}
\includegraphics[width=0.75\textwidth]{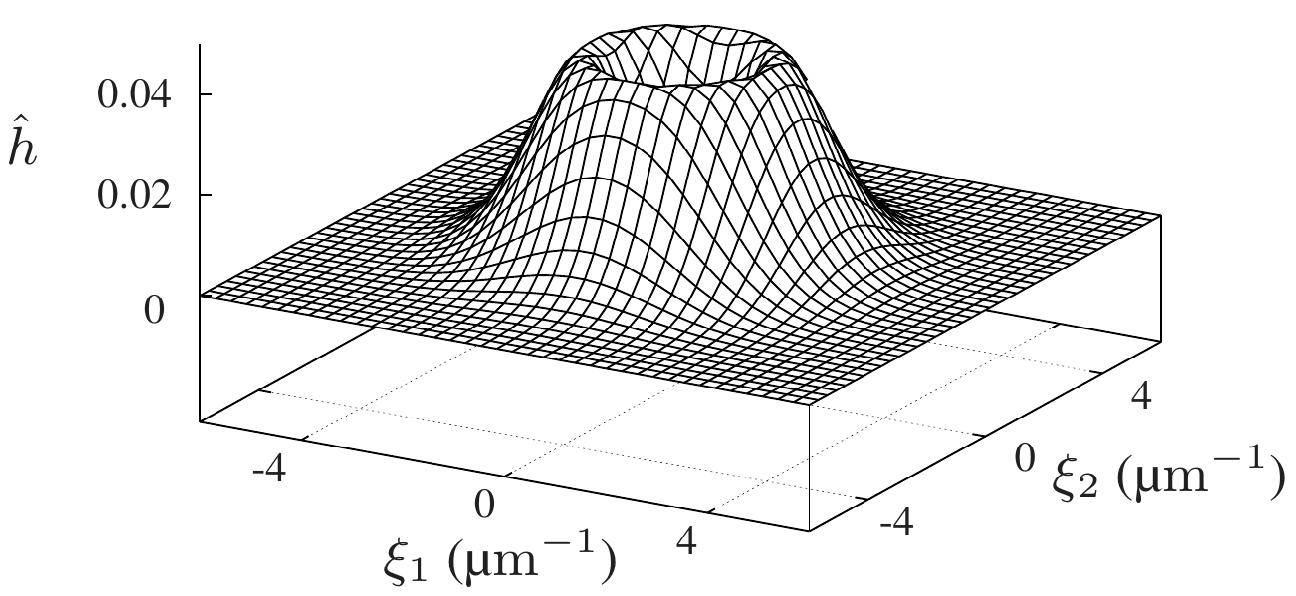}
\end{center}
\caption{Transfer function $\hat{h}(\xi)$ of the DoG filter $h(x)$
  with $\sigma_1=\frac{1}{\sqrt{2}}\,$\textmu m and $\sigma_2=1\,$\textmu m.}
\label{figDoG2}
\end{figure}

\begin{figure}[htb]
\begin{center}
\includegraphics[width=0.75\textwidth]{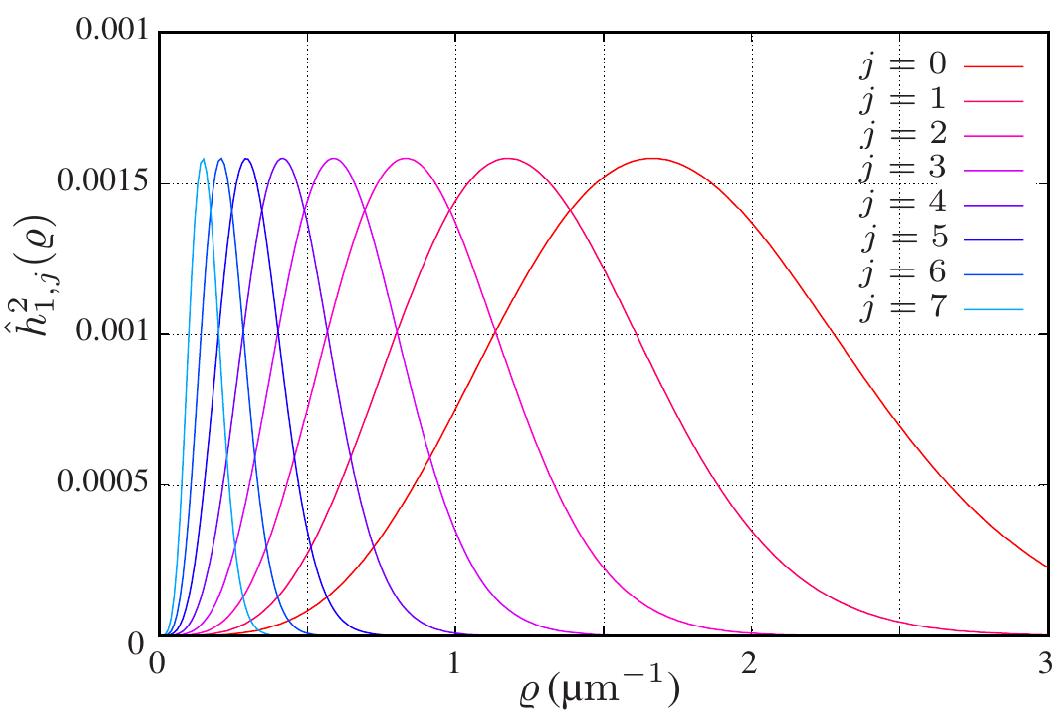}
\end{center}
\caption{The square $\hat{h}_{1,j}^2(\varrho)$ of the transfer function of the DoG filter for $j=0,\ldots, 7$.}
\label{figDoG3}
\end{figure}

We use Plancherel's equation to shed light on the close relationship of the Laplace
pyramid and the power spectrum.  Plancherel's equation says that for
all functions of fast decay, the norm of the function is the same as the norm of its
Fourier transform \cite{katznelson04}. 
Together with the Fourier convolution theorem this yields
\[ {\rm MSP}\!_j \approx \frac{2\pi}{A}\int\limits_{\R^2}\bigl|\hat{f}_W(\xi)\cdot\hat{h}_j(\xi)\bigr|^2\,dx,\]
where the Fourier transform
\[ \hat{h}_j(\xi) = 
\frac{1}{2\pi}\Bigl(\exp(-\sigma_{j-1}^2\|\xi\|^2)-\exp(-\sigma_j^2\|\xi\|^2)\Bigr),\qquad\xi\in\R^2,\]
of the mask $\hat{h}_j(x)$ is known as the transfer function of the
DoG filter. See Figure~\ref{figDoG2} for an example of
$\hat{h}_j(\xi)$. Define the radial function $\hat{h}_{1,j}^2(\varrho)$ of
$\hat{h}_j^2(\xi)$ by $\hat{h}_{1,j}^2(\|\xi\|)=\hat{h}_j^2(\xi)$ for
$\xi\in\R^2$, see Figure~\ref{figDoG3}. The mask $\hat{h}_j(\xi)$ is positive for all $\xi$. Consequently,
\begin{eqnarray*} 
{\rm MSP}\!_j &\approx& \frac{2\pi}{A}\int\limits_{\R^2}\bigl|\hat{f}_W(\xi)\bigr|^2\,\hat{h}_j^2(\xi)\,d\xi\\
&=&\int\limits_{\R^2}\hat{k}(\xi)\,\hat{h}_j^2(\xi)\,d\xi
= 2\pi\int\limits_0^\infty\varrho\,\hat{k}_1(\varrho)\,\hat{h}_{1,j}^2(\varrho)\,d\varrho.
\end{eqnarray*}
This means, the square of the transfer function of the filter cuts a frequency band from the power spectrum of
$f(x)$. Subsequently, the total power is calculated in this band. 
The transfer function $\hat{h}_{1,j}(\varrho)$ attains its maximum at
$\varrho_{\rm max}=1.177/\sigma_j$. 
The square of the norm of $\hat{h}_{1,j}(\varrho)$ is
\[ \|\hat{h}_j\|^2 
=\int\limits_{\R^2} |\hat{h}_j(\xi)|^2\,d\xi 
= 2\pi\int\limits_0^\infty \varrho\,\hat{h}_{1,j}^2(\varrho)\,d\varrho
=\frac{1}{24\pi\sigma_j^2}.\]

For different $j$, the functions $\hat{h}_{1,j}^2(\varrho)$ differ 
considerably in the full width at half maximum and they overlap strongly. 
Therefore, the ${\rm MSP}\!_j$ do not contain independent information
about the spectrum of the nonwoven. Changing the variances 
$\sigma_j^2$ and $\sigma_{j+1}^2$ of consecutive Gaussian functions in
$h_j(x)$ such that their difference is smaller, does not remedy this. 
Table~\ref{tabMSP} yields the estimated ${\rm MSP}\!_j$ for
indices $j$ with position $\varrho_{\rm max}$ of the maximum
transfer function $\hat{h}_{1,j}(\varrho)$ in the frequency band
$[\varrho_0,\varrho_1]$ chosen for estimating the cloudiness
index ${\rm CLI}$. 
The quantities ${\rm MSP}\!_j$ and $\hat{k}_1(\varrho)$ are completely
different by definition. However, they are actually comparable in the
following sense: The ratio $2\pi{\rm MSP}\!_j/\|\hat{h}_j\|^2$ is in
the same order of magnitude as $\hat{k}_1(\varrho_{\rm max})$. For
example, for the nonwoven FL1 we have
$2\pi{\rm MSP}\!_9/\|\hat{h}_9\|^2=26.3\,$\textmu m$^2$ and
$\hat{k}_1(\varrho_{\rm max})=25.7\,$\textmu m$^2$.

\begin{figure}[htb]
\begin{center}
\includegraphics[width=0.75\textwidth]{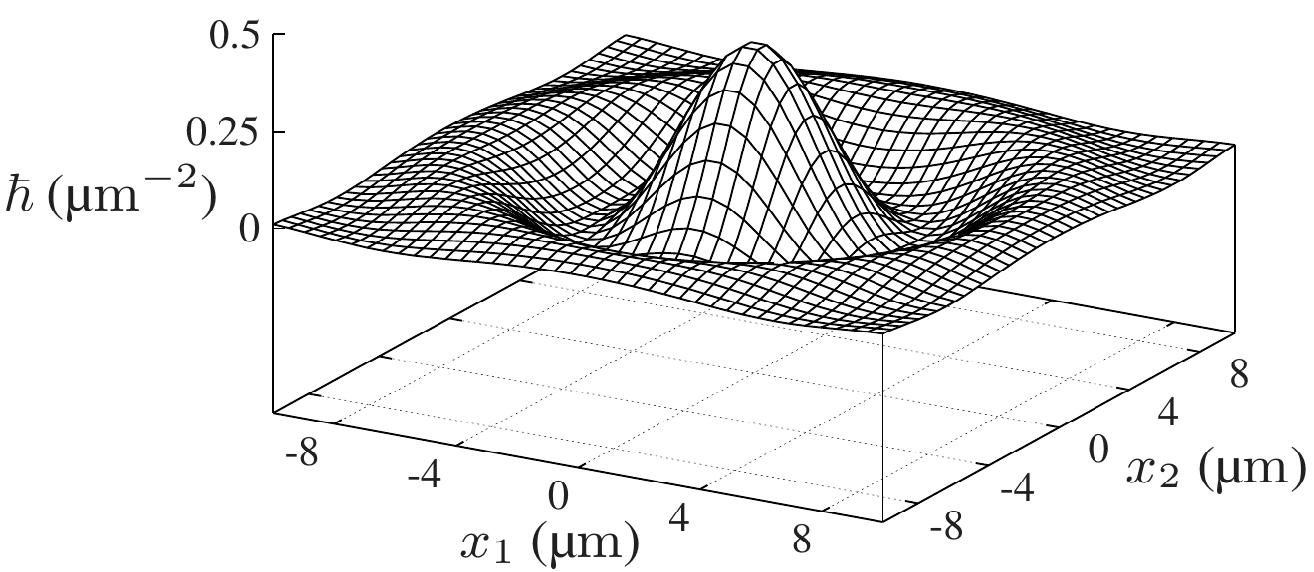}
\end{center}
\caption{Filter mask $\hbar(x)$ of the Bessel band pass filter
  with frequency bounds $\varrho_0=\frac{1}{2}\,$\textmu m$^{-1}$ 
  and $\varrho_1=1\,$\textmu m$^{-1}$.}
\label{figDoG5}
\end{figure}

Finally, note that a spectral band could be separated sharply by the filter mask
\[ \hbar(x)=
\frac{\varrho_1\,J_1(\varrho_1\|x\|)}{\|x\|}-\frac{\varrho_0\,J_1(\varrho_0\|x\|)}{\|x\|},
\qquad x\in\R^2.\]
The function $\hbar(x)$ is known as the mask of the Bessel band pass
filter, see Figure~\ref{figDoG5} for an example. Its transfer function
is the indicator function of the circular ring with 
radii $\varrho_0$ and $\varrho_1$, i.\,e.\ 
$\hat{\hbar}(\xi)=\ind_{B_{\varrho_1}}(\xi)-\ind_{B_{\varrho_0}}(\xi)$
for $\xi\in\R^2$, where $\varrho_0$ and $\varrho_1$ act like 
sharp frequency bounds. The result would be equivalent to
evaluating cloudiness based on our CLI from \eqref{eqCLI}.

\begin{table}[htb]
\caption{Estimates of ${\rm MSP}\!_j$ for $j=9,\ldots,12$. 
The error is the standard deviation of the mean of the esti\-mates 
from the in\-di\-vi\-du\-al images. 
The position $\varrho_{\rm max}$ of the maximum and the norm
$\|\hat{h}_j\|$ of the trans\-fer function $\hat{h}_j(\xi)$ are
included to facilitate interpretation.}
\label{tabMSP} 
\begin{center}
{
\renewcommand{\arraystretch}{1.2}
\begin{tabular}{|c|c||r|r|r|r|}
\hline
spe- & $m$ & \multicolumn{4}{c|}{${\rm MSP}\!_j$ (\permil)} \\\cline{3-6}
cimen & & {\small $j=9$} & {\small $j=10$} & {\small $j=11$} & {\small $j=12$} \\
\hline\hline
FL1 & 14 & $0.236$ & $0.272$ & $0.307$ & $0.338$ \\ 
    & & $\pm0.007$   & $\pm0.007$ & $\pm0.007$ & $\pm0.008$\\\hline 
FL2 & 10 & $0.294$ & $0.326$ & $0.349$ & $0.353$ \\ 
    & & $\pm0.004$ & $\pm0.004$ & $\pm0.004$ & $\pm0.005$ \\\hline 
FL3 & 10 & $0.331$ & $0.363$ & $0.377$ & $0.364$ \\ 
    & & $\pm0.003$ & $\pm0.004$ & $\pm0.004$ & $\pm0.004$ \\ 
\hline\hline
\multicolumn{2}{|c||}{ ~} & ~ & ~ & ~ &\vspace{-0.35 cm}\\
\multicolumn{2}{|c||}{$\varrho_{\rm max}\,$(\textmu m$^{-1}$)} & $0.074$ & $0.052$ & $0.037$ & $0.026$ \\\hline
\multicolumn{2}{|c||}{ ~} & ~ & ~ & ~ &\vspace{-0.35 cm}\\
\multicolumn{2}{|c||}{\!\!$\|\hat{h}_j\|\,$(mm$^{-1}$)} & $7.200$ & $5.090 $ & $3.600$ & $2.545$ \\
\hline
\end{tabular}
}
\end{center}
\end{table}

\section{Discussion and conclusion}

We define the cloudiness of nonwovens as the total power of the
normalized local grammage $f(x)$ in a frequency band. Our cloudiness
index ${\rm CLI}$ is therefore independent of the nominal grammage
$\bar{w}$ and the standard deviation $\sigma_w$ of the local grammage
$w(x)$. Estimating $\bar{w}$ and $\sigma_w$ requires careful optical
densitometry measurements, whereas estimating ${\rm CLI}$ does
not. ${\rm CLI}$ thus simplifies estimating cloudiness based on
light-optical transmission images considerably. Moreover, the
associated technical effort is low. As a consequence,
${\rm CLI}$ is very attractive for quality control in industry.

The power spectrum determined by image analysis is closely related to
other cloudiness measures, as discussed in
Section~\ref{sec:other-measures}: The range of interaction is -- up to
the factor $2\pi$ -- nothing but the power spectrum at the origin
(Section~\ref{sec:range-interaction}).  Physical methods (SAXS, sSAXS)
yield the scattering intensity as a function of the
magnitude of the momentum transfer scattering vector. This 
corresponds -- up to a multiplicative
constant -- to the power spectrum (Section~\ref{sec:small-angle-scatt}). 
Finally, the normalized mean square of the pixel values of a
transmission image processed with a DoG filter (Laplacian pyramid)
is the power spectrum at a fixed frequency 
(Section~\ref{sec:laplace-pyramid}). Hence, all
these methods intuitively amount to the same and obtaining 
comparable cloudiness values just requires careful
adjustment of the respective parameters.

The definition
of the power spectrum is tied to the definition of the continuous
Fourier transform. Commonly used definitions of the Fourier transform
differ in the use of the normalization, the sign of the exponents, and
using circular frequency instead of frequency.
Another source of deviation is the step from continuous to discrete
Fourier transform. Here, we understand the discrete FFT as a
numerical version of the continuous Fourier transform, see e.\,g.\
\cite[page 607ff]{press07}. 
Care is needed to estimate the power spectrum
based on Equation~\eqref{eqEstPow} reproducibly, in particular with respect to
normalization as various versions of the FFT are
used in the literature and in software packages. 

\section*{Declarations}
\subsection*{List of abbreviations}
\begin{description}
\item{FFT} fast Fourier transform
\item{SAS} small-angle scattering
\item{SAXS} small-angle X-ray scattering 
\item{sSAXS} scanning small-angle X-ray scattering 
\item{DoG} difference of Gaussian (filter)
\end{description}
 
\subsection*{Availability of data and materials}

The images used in this paper along with a ToolIP \cite{toolip} graph 
estimating the CLI will be provided on
\url{https://www.itwm.fraunhofer.de/cloudiness}. 

\subsection*{Funding}

This research was supported by the project ``Product and quality
optimization for protective clothing against infections made from nonwovens'' 
(ProQuIV) within the Fraunhofer Society's anti-Corona programme. 

\subsection*{Acknowledgements}

We thank ITWM colleagues Ralf Kirsch and Kai Taeubner for providing the
samples and their properties and for the microscopic imaging, respectively. 

\begin{backmatter}

\bibliographystyle{bmc-mathphys}
\bibliography{References,Paper,Tomography}

\end{backmatter}
\end{document}